\documentclass[sigconf]{acmart}

\AtBeginDocument{%
  }

\copyrightyear{2025}
\acmYear{2025}
\setcopyright{acmlicensed}\acmConference[WWW '25]{Proceedings of the ACM Web Conference 2025}{April 28-May 2, 2025}{Sydney, NSW, Australia}
\acmBooktitle{Proceedings of the ACM Web Conference 2025 (WWW '25), April 28-May 2, 2025, Sydney, NSW, Australia}
\acmDOI{10.1145/3696410.3714905}
\acmISBN{979-8-4007-1274-6/25/04}

\usepackage{booktabs}
\usepackage{graphicx}
\usepackage{multirow}
\usepackage{subfigure}
\usepackage{verbatim}
\usepackage{amsmath}
\usepackage{caption}
\usepackage[normalem]{ulem}
\useunder{\uline}{\ul}{}
\usepackage{algorithm}
\usepackage{algorithmic}

\theoremstyle{plain}
\newtheorem{theorem}{Theorem}[section]

\newtheorem{corollary}[theorem]{Corollary}
\theoremstyle{definition}

\theoremstyle{remark}

\begin{document}

\title{Dual-level Mixup for Graph Few-shot Learning with Fewer Tasks}
%\title{Dual-level Mixup for Few-shot Node Classification \\ with Fewer Tasks}



\author{Yonghao Liu}
\authornote{Key Laboratory of Symbolic Computation and Knowledge Engineering of the Ministry of Education}
\author{Mengyu Li}
\authornotemark[1]
\affiliation{
  \institution{College of Computer Science and Technology, Jilin University}
  \city{Changchun}
  \country{China}
}
\email{yonghao20@mails.jlu.edu.cn}
\email{mengyul21@mails.jlu.edu.cn}

% \author{Mengyu Li}
% \affiliation{
%   \institution{College of Computer Science and Technology, Jilin University}
%   \city{Changchun}
%   \country{China}
% }
% \email{mengyul21@mails.jlu.edu.cn}

\author{Fausto Giunchiglia}
\affiliation{%
  \institution{Department of Information Engineering and Computer Science, University of Trento}
  \city{Trento}
  \country{Italy}
}
\email{fausto.giunchiglia@unitn.it}

\author{Lan Huang}
\authornotemark[1]
\affiliation{%
  \institution{College of Computer Science and Technology, Jilin University}
  %\streetaddress{}
  \city{Changchun}
  %\state{}
  \country{China}
}
\email{huanglan@jlu.edu.cn}

\author{Ximing Li}
\authornotemark[1]
\affiliation{%
  \institution{College of Computer Science and Technology, Jilin University}
  \city{Changchun}
  \country{China}
}
\email{liximing86@gmail.com}

\author{Xiaoyue Feng}
\authornotemark[1]
\authornote{Corresponding author}
\affiliation{%
  \institution{College of Computer Science and Technology, Jilin University}
  \city{Changchun}
  \country{China}
}
\email{fengxy@jlu.edu.cn}

\author{Renchu Guan}
\authornotemark[1]
\authornotemark[2]
\affiliation{%
  \institution{College of Computer Science and Technology, Jilin University}
  \city{Changchun}
  \country{China}
}
\email{guanrenchu@jlu.edu.cn}

\renewcommand{\shortauthors}{Yonghao Liu et al.}
\begin{abstract}
Graph neural networks have been demonstrated as a powerful paradigm for effectively learning graph-structured data on the \textit{web} and mining content from it. %the \textit{wide web}. for downstream task analysis. 
Current leading graph models require a large number of labeled samples for training, which unavoidably leads to overfitting in few-shot scenarios. Recent research has sought to alleviate this issue by simultaneously leveraging graph learning and meta-learning paradigms. However, these graph meta-learning models assume the availability of numerous meta-training tasks to learn transferable meta-knowledge. Such assumption may not be feasible in the real world due to the difficulty of constructing tasks and the substantial costs involved. Therefore, we propose a \textbf{S}i\textbf{M}ple yet effect\textbf{I}ve approach for graph few-shot \textbf{L}earning with f\textbf{E}wer tasks, named \textbf{SMILE}. We introduce a dual-level mixup strategy, encompassing both within-task and across-task mixup, to simultaneously enrich the available nodes and tasks in meta-learning. Moreover, we explicitly leverage the prior information provided by the node degrees in the graph to encode expressive node representations. Theoretically, we demonstrate that SMILE can enhance the model generalization ability. Empirically, SMILE consistently outperforms other competitive models by a large margin across all evaluated datasets with in-domain and cross-domain settings. Our anonymous code can be found \href{https://github.com/KEAML-JLU/SMILE}{\textcolor{red}{here}}.
\end{abstract}

\begin{CCSXML}
<ccs2012>
   <concept>
       <concept_id>10002951.10003227.10003351</concept_id>
       <concept_desc>Information systems~Data mining</concept_desc>
       <concept_significance>500</concept_significance>
       </concept>
   <concept>
       <concept_id>10010147.10010257.10010293.10010294</concept_id>
       <concept_desc>Computing methodologies~Neural networks</concept_desc>
       <concept_significance>500</concept_significance>
       </concept>
 </ccs2012>
\end{CCSXML}

\ccsdesc[500]{Information systems~Data mining}
\ccsdesc[500]{Computing methodologies~Neural networks}

\keywords{Graph neural network, Few-shot learning, Node classification}

\maketitle

\section{Introduction}
As a fundamental data structure, graphs can effectively model complex relationships between objects and they are ubiquitous in the real world. %Node classification, as a crucial task in graph learning, has consistently garnered sustained attention from researchers. 
Graph neural networks (GNNs) %, due to their powerful representation capabilities for graph-structured data, 
have been widely employed as an effective tool for graph %to learn informative node embeddings for downstream 
task analysis \cite{kipf2016semi, liu2021vplag, liu2021deep, liu2023time, liulocal, li2024simple, liu2024simple, liu2024improved, liu2024resolving, liu2025boosting, liu2025improved, liu2025enhancing}. Prevailing GNN models are designed under the supervised learning paradigm, which implies that they require abundant labeled data to achieve satisfactory classification performance \cite{ding2020graph, tan2022transductive}. Given the limited number of labeled nodes per class, known as few-shot cases \cite{zhang2022few, liu2022few}, 
these models suffer from severe overfitting, leading to a significant performance decline \cite{liu2019learning, huang2020graph}. %In cases of scarce labels, particularly in few-shot scenarios, they are highly susceptible to overfitting issues, leading to severe performance degradation \cite{liu2019learning}. 

Meta-learning has emerged as a viable option for effectively learning from limited labeled data. %The currently prevalent meta-learning models, 
Its core concept is to train on tasks instead of instances as training units, aiming to capture the differences between tasks to enhance the model generalizability \cite{hospedales2021meta}. Several pioneering models \cite{zhou2019meta, wang2022task} have attempted to leverage integrate GNNs and meta-learning techniques to address graph few-shot learning problems. %few-shot node classification, a representative task in graph few-shot learning, %with few-shot node classification as a representative task, and have achieved remarkable performance. %with impressive performance. 
However, these graph meta-learning models all assume the existence of abundant accessible meta-training tasks to extract generalizable meta-knowledge for rapid adaptation to meta-testing tasks with only a few labeled instances. In other words, their outstanding performance critically depends on a wide range of meta-training tasks. %all presuppose the availability of a substantial number of accessible meta-training tasks, which are utilized to extract generalizable meta-knowledge for swift adaptation to meta-testing tasks characterized by a limited number of labeled instances. In essence, the exceptional performance of these models is heavily contingent on a diverse array of meta-training tasks.
For many real-world applications, due to the difficulty of task generation or data collection, we may not be able to obtain an adequate number of meta-training tasks \cite{tan2022transductive, yao2021meta, lee2022set}. For molecular property prediction, %given the limited known chemical properties (\textit{i.e.}, classes), 
labeling newly discovered chemical compounds requires extensive domain knowledge and expensive wet-lab experiments \cite{Guo21few}. Moreover, even after annotation, the currently known chemical properties (\textit{i.e.}, classes) are limited, encompassing only common molecular characteristics such as polarity, solubility, and toxicity \cite{livingstone2000characterization}. %Therefore, it is infeasible to construct numerous meta-training tasks in the real-world. %For example, 
%If we only know six types of chemical properties, each with a few labeled compounds, then we can construct at most six 5-way meta-training tasks. %For example, in biological networks, numerous domain knowledge is required to label newly discovered protein nodes, which is a challenging task even for veteran researchers \cite{hu2019strategies}. 

\begin{figure}
    \centering
    \subfigure[Amazon-Clothing]{\includegraphics[width=0.22\textwidth]{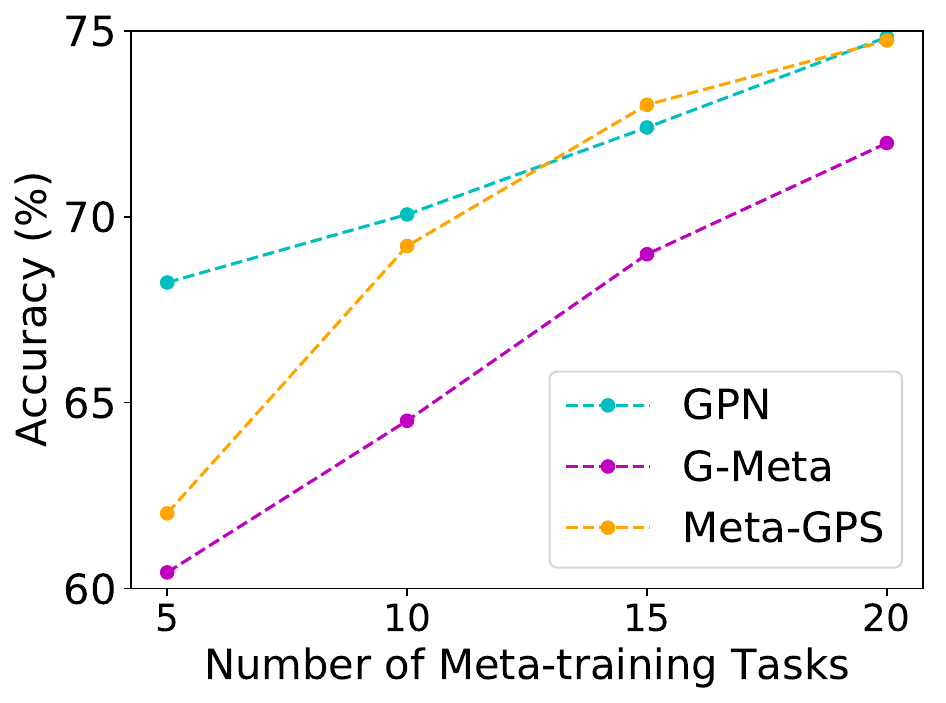}}
        \subfigure[Cora-Full]{\includegraphics[width=0.22\textwidth]{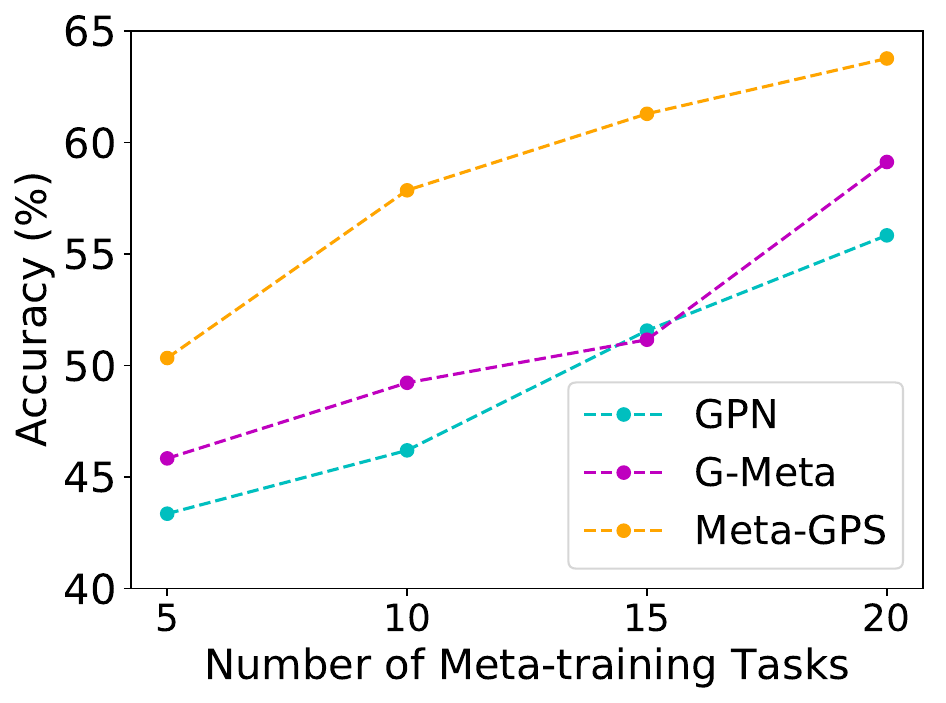}}
    \caption{Model performance varies with the number of meta-training tasks across different datasets.}
    \label{comparison}
\end{figure}

To further support our argument, we select three representative graph meta-learning models (\textit{i.e.}, GPN \cite{ding2020graph}, G-Meta \cite{huang2020graph}, and Meta-GPS \cite{liu2022few}) and evaluate their performance under varying numbers of meta-training tasks in Fig. \ref{comparison}. We distinctly observe that as the number of available meta-training tasks decreases, the overall performance of all methods greatly deteriorates. Because they tend to memorize meta-training tasks directly, which significantly constrains their generalization ability to novel tasks in the meta-testing stage \cite{rajendran2020meta}.
%\vspace{-0.5em}
%\vspace{-0.01em}
This naturally raises a pressing question for us in more realistic scenarios: \textit{How can we perform graph few-shot learning in scenarios with fewer tasks to extract as much transferable meta-knowledge as possible, thereby enhancing the model generalization performance?} Regarding this, although some recent studies \cite{tan2023virtual, kim2023task} have made some efforts on this issue, they primarily employ intricate network architectures to endow models with favorable characteristics, yet there is still room for improvement. We argue that there are two serious issues for our focused scenarios, which greatly hamper the model performance. \textit{On the one hand}, there are only limited support samples available for training within each meta-training task, which complicates the accurate reflection of the real data distribution. Therefore, it poses a challenge for the model to effectively capture the data characteristics, severely affecting its inductive bias capability \cite{ni2021data}. \textit{On the other hand}, when there are only limited meta-training tasks available, the model tends to directly fit the biased task distribution. This implies not only a shortage of data for each task, but also a reduced number of available tasks. %This means that not only is there a scarcity of data for each task, but there are also fewer tasks available. %It means that not only is there scarcity of data in each task, but also fewer tasks available. %That be saying, not only is there scarcity of data in each task, but there are also fewer tasks available. %In summary, there is a scarcity of data within each task, and there are also fewer tasks available. 
The combined effect of these two factors increases the unnecessary oscillation during predictions outside the training examples, leading to reduced generalization capability. %These two factors together increase the unnecessary oscillation of the model when making predictions outside the training examples, leading to reduced generalization capability.
%As the number of available meta-training tasks decreases, these models are susceptible to overfitting as they tend to memorize ones directly, significantly constraining their generalization ability to novel tasks in the meta-testing stage. We select several representative graph meta-learning models, and show their performance with varying numbers of meta-training tasks in Fig. \ref{comparison}. The results distinctly support our argument. %The results presented in Fig. \ref{comparison} distinctly corroborates our argument.

%However, optimizing the single model is prone to overfitting since we are given only a small number of meta-training tasks. The meta-learner tends to memorize the meta-training tasks, which limits its generalization to new tasks at meta-test time \cite{ yin2019meta, rajendran2020meta}.
%Our model considerably outperforms other baselines and can even achieve up to 20\% improvement on certain datasets.

To address these issues mentioned above, we develop a \textbf{S}i\textbf{M}ple yet effect\textbf{I}ve approach for graph few-shot \textbf{L}earning with f\textbf{E}wer tasks, namely, \textbf{SMILE}. %In this work, we mainly focus on few-shot node classification tasks, which is a fundamental task extensively studies by previous researches. 
Specifically, given the graph data, we first obtain discriminative node embeddings using our designed graph encoder. In this process, we introduce node degrees as prior information to fully utilize valuable information present in the existing graph. Then, we introduce a dual-level mixup strategy that operates on the obtained hidden node representations, consisting of both within-task and across-task mixup. The former involves random sampling of two instances from the same category within a task and applies linear interpolation to generate new samples, thereby enriching the data distribution. The latter requires computing class prototypes in two randomly selected original meta-training tasks, and then linearly interpolating class prototypes from different tasks to generate new tasks, thereby densifying the task distribution. %The latter entails computing class prototypes within randomly selected two original meta-training tasks, and then applying linear interpolation to classes prototypes from different tasks to generate new tasks, thereby densifying the task distribution. 
These two employed strategies effectively mitigate the adverse effects caused by sample and task scarcity. %The two strategies employed effectively mitigate the adverse effects brought about by sample scarcity and task scarcity. %The former randomly samples two instances of the same category within a task and performs linear interpolation to generate new instances, enriching the data distribution. While the latter randomly samples two original meta-training tasks and performs linear interpolation to generate new tasks, densifying the task distribution.
Empirically, despite its simplicity and the absence of sophisticated techniques, the proposed approach demonstrates remarkable performance. %exhibits impressive performance. 
Furthermore, we provide a theoretical elucidation of the underlying mechanism of our method, demonstrating its ability to constrict the upper bound of generalization error and consequently achieve superior generalization. %Moreover, theoretically, we elucidate the underlying mechanism of our method, showcasing its capacity to tight the upper bound of generalization error and thus achieve superior generalization. %we present the working mechanism behind our method, demonstrating its ability to tight the generalization error, thereby achieving better generalization. 
In summary, our contributions can be summarized as follows:

\noindent $\bullet$ We propose a simple yet effective approach, SMILE, which leverages dual-level task mixup technique and incorporates the node degrees prior information, for graph few-shot learning with fewer tasks. 

\noindent $\bullet$ We theoretically analyze the reasons why our approach works, demonstrating its ability to enhance generalization performance by regularizing model weights.

\noindent $\bullet$ We conduct extensive experiments on the several datasets, and the results show that SMILE can considerably outperform other competitive baselines by a large margin with in-domain and cross-domain settings.
\section{Related Work}
\subsection{Few-shot Learning}
%\paragraph{Few-shot Learning.}
%\noindent \textbf{Few-shot Learning.} 
Few-shot learning aims to quickly adapt meta-knowledge acquired from previous tasks to novel tasks with only a small number of labeled samples, thereby enabling few-shot generalization of machine learning algorithms \cite{finn2017model, finn2019online, hospedales2021meta}. %Initially, it was widely applied in the fields of computer vision and natural language processing to address real-world problems. 
Typically, there are three main strategies to solve few-shot learning. Some methods \cite{vinyals2016matching, snell2017prototypical, mishra2017simple} utilize prior knowledge to constrain the hypothesis space at the \textit{model level}, learning a reliable model within the resulting smaller hypothesis space. A series of methods \cite{finn2018probabilistic, flennerhag2019meta, grant2018recasting, rusu2018meta, oh2020boil} improve the search strategy at the \textit{algorithm level} by providing good initialization or guiding the search steps. Another line of works \cite{hariharan2017low, xu2021augnlg, yao2021improving} augment tasks at the \textit{data level} to obtain precise empirical risk minimizers. For the few-shot learning with limited tasks, there are various explorations in Euclidean data, such as images and texts. For example, TAML \cite{jamal2019task} and MR-MAML \cite{yin2019meta} directly apply regularization on the few-shot learner to reduce their reliance on the number of tasks. Meta-aug \cite{rajendran2020meta} and MetaMix \cite{yao2021improving} perform data augmentation on individual tasks to enrich the data distribution. MLTI \cite{yao2021meta} and Meta-Inter \cite{lee2022set} directly generate source tasks to densify the task distribution. 
%MetaMix linearly combines features and labels from samples in the support and query sets to augment meta-training tasks with more data. MLTI proposes generating new tasks through interpolation between existing meta-training tasks to alleviate the problem of insufficient tasks. Modulation uses neural networks to adjust batch normalization parameters during meta-training to increase task density, while modifying parameters at various network levels to enhance task diversity.

%Our approach differs from the aforementioned methods in that we simultaneously perform within-task and across-task interpolation, with each strategy playing a crucial role. Meanwhile, we integrated task prototypes into the mixup process and explicitly adopt those previously overlooked original tasks, resulting in superior performance, which can be supported by the results in \textbf{Appendices} \ref{alternative_mixup} and \ref{ori_task}. Moreover, previous methods are not applicable to graph-structured data, whereas SMILE introduces a strategy that leverages the prior information provided by the graph. %Additionally, while those methods are not applicable to graph-structured data, we directly leverage the prior information in the graph by considering node degrees.

\subsection{Graph Few-shot Learning}
%\paragraph{Graph Few-shot Learning.}
%\noindent \textbf{Graph Few-shot Learning.}
Inspired by the success of few-shot learning in computer vision \cite{lifchitz2019dense, bateni2020improved, tian2020rethinking} and natural language processing \cite{mukherjee2020uncertainty, bao2019few, wang2021grad2task}, few-shot learning on graphs has recently seen significant development \cite{liu2022few, wang2022task, wang2023few, tan2023virtual, liu2024meta, liu2024simple, liu2025enhancing}. %Few-shot node classification is a fundamental and widely popular task aimed at training a graph model to rapidly identify novel classes given a limited number of labeled nodes. %The task of few-shot node classification is the most popular, aiming to train a graph model to rapidly identify new classes given a limited number of labeled nodes.
The core concept of current mainstream methods is to develop complicated algorithms to address the problem of few-shot learning on graphs. For instance, Meta-GNN \cite{zhou2019meta}, G-Meta \cite{huang2020graph}, and Meta-GPS \cite{liu2022few} are all subjected to specific modifications based on the MAML \cite{finn2017model} algorithm, employing a bi-level optimization strategy to learn better parameter initialization. %Despite fruitful success, the high complexity of these models severely hinder their further improvements. 
While the above models yield fruitful results, their reliance on substantial and diverse of meta-training tasks, coupled with their high complexity, has impeded their further advancement. %Additionally, these models have not altered the data distribution, and there
Recently, TLP \cite{tan2022transductive} and TEG \cite{kim2023task} attempt to alleviate the limited diversity in meta-training datasets by using graph contrastive learning and equivariant neural networks, respectively. %Moreover, some proposed graph prompting methods \cite{liu2023graphprompt, fang2024universal}, by harnessing the power of prompting techniques, have emerged as effective methods under the scarcity of labeled nodes. 
With the aid of sophisticated network designs, these methods have yielded promising results in graph few-shot scenarios. %Benefiting from sophisticated network designs, the above methods have achieved promising results in graph few-shot scenarios. 
However, there is little effort to address the graph few-shot learning problem from the perspective of data augmentation.
\section{Method}
\begin{figure*}
    \centering
    \includegraphics[width=0.8\textwidth]{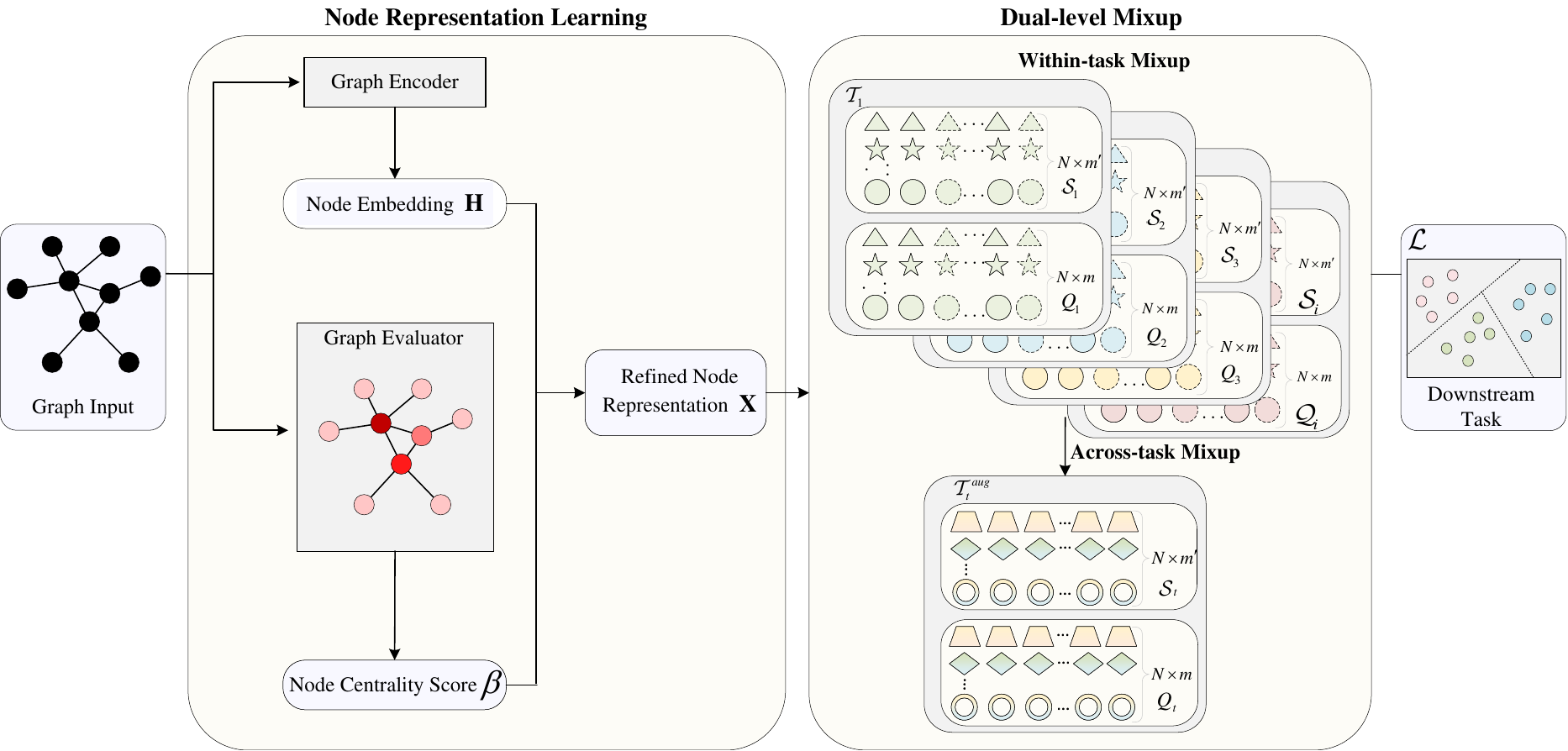}
    \caption{The overall architecture of SMILE.}
    \label{overall}
\end{figure*}

In this section, we first present the preliminary knowledge. Then, we detail our proposed SMILE, which consists of two components: node representation learning and dual-level mixup strategy. To facilitate better understanding, we present the overall framework of the model in Fig. \ref{overall}.
%\vspace{-1em}

\subsection{Preliminary}
Given a graph $\mathcal{G}\!=\!\{\mathcal{V},\mathcal{E},\mathrm{Z},\mathrm{A}\}$, $\mathcal{V}$ and $\mathcal{E}$ represent the sets of nodes and edges, respectively. $\mathrm{Z}\!\in\!\mathbb R^{n\times d}$ is the feature matrix of nodes and $\mathrm{A}\!\in\!\mathbb R^{n\times n}$ is the corresponding adjacency matrix. Our model adheres to the prevalent meta-learning training paradigm, which involves training on sampled tasks. In this work, we mainly focus on few-shot node classification, which is the most prevailing and representative task in graph few-shot learning. Moreover, we highlight that in our focused scenarios, the number of available meta-training tasks sampled from an unknown task distribution is extremely small compared to traditional experimental settings, referred to as \textit{few-shot node classification with fewer tasks}. Our goal is to enable the model to effectively extract meta-knowledge even from such limited tasks, which can generalize to novel tasks in the meta-testing phase. For better understanding, we summarize the main symbols of this work in \textbf{Appendix} \ref{symbol}.

\subsection{Node Representation Learning}

%\noindent \textbf{Node Representation Learning.}
Generally, the initial step involves encoding the nodes within the graph into a latent space, thereby transforming them into low-dimensional hidden vectors. GNNs have become the foremost choice for node embedding due to its powerful representational capabilities on graphs. It follows a message-passing mechanism, continuously aggregating messages from neighboring nodes to iteratively update the embedding of the target node. Guided by the simple philosophy, we adopt the SGC model \cite{wu2019simplifying} to learn node embeddings. Specifically, which can be defined as:
\begin{equation}
\label{sgc}
\begin{aligned}
\mathrm{H}\!=\!\breve{\mathrm{A}}\cdots\breve{\mathrm{A}}\mathrm{Z}\mathrm{W}^{(0)}\mathrm{W}^{(1)}\cdots\mathrm{W}^{(\ell-1)}\!=\!\breve{\mathrm{A}}^\ell\mathrm{Z}\mathrm{W}^\ast,
\end{aligned}   %\mathrm{H}^{\ell+1}\!=\!\breve{\mathrm{A}}\mathrm{H}^{\ell},
\end{equation}
%where $\breve{\mathrm{A}}\!=\!\hat{\mathrm{D}}^{-1/2}\hat{\mathrm{A}}\hat{\mathrm{D}}^{-1/2}$ is the symmetric normalized adjacency matrix with added self-loops, \textit{i.e.}, $\hat{\mathrm{A}}\!=\!\mathrm{A}+\mathrm{I}$. $\hat{\mathrm{D}}_i\!=\!\sum_j\hat{\mathrm{A}}_{i,j}$ denotes the corresponding degree matrix. $\mathrm{H}^{\ell}$ is the $\ell$-th node embeddings and $\mathrm{H}^{0}\!=\!\mathrm{Z}$ is the initialized node features. After performing graph convolution operations, we can obtain the node vectors $\mathrm{H}\!\in\!\mathbb R^{n\times d}$ that simultaneously encode node features and topology structure. %, denote as $\mathrm{X}$. %as the output node embeddings by the graph encoder.
where $\breve{\mathrm{A}}\!=\!\hat{\mathrm{D}}^{-1/2}\hat{\mathrm{A}}\hat{\mathrm{D}}^{-1/2}$ is the symmetric normalized adjacency matrix with added self-loops, \textit{i.e.}, $\hat{\mathrm{A}}\!=\!\mathrm{A}+\mathrm{I}$. $\hat{\mathrm{D}}_i\!=\!\sum_j\hat{\mathrm{A}}_{i,j}$ denotes the corresponding degree matrix. $\mathrm{W}^{\ast}$ is the collapsing weight matrices. After performing graph convolution operations, we can obtain the node vectors $\mathrm{H}\!\in\!\mathbb R^{n\times d}$ that simultaneously encode node features and topology structure. %, denote as $\mathrm{X}$. %as the output node embeddings by the graph encoder.

Given that few-shot models are highly noise-sensitivity \cite{zhang2019variational}, it is necessary to incorporate more prior knowledge to refine representations. Such prior knowledge is often reflected on node degree about the node popularity and importance \cite{Park2019estimating}. %Therefore, we consider explicitly incorporating it into the node encoding process. Specifically, we first perform linear transformation to the derived node features to assign initial scores $\kappa\!\in\!\mathbb R^{n\times 1}$. Then, based on the node degree information, we obtain the node centralities $\alpha\!\in\!\mathbb R^{n\times 1}$ to adjust the initial scores. Finally, we acquire the refined node representations $\mathrm{X}\!\in\!\mathbb R^{n\times d}$ using the adjusted scores $\beta\!\in\!\mathbb R^{n\times 1}$. The above procedures can be formulated as follows:
Therefore, we consider explicitly incorporating it to evaluate each node. Specifically, we first adopt another SGC to derive the interaction weights $\kappa\!\in\!\mathbb R^{n\times 1}$ for all nodes. Then, based on the node degree information, we obtain the node centralities $\alpha\!\in\!\mathbb R^{n\times 1}$ to perform degree normalization for adjusting $\kappa$. Finally, we acquire the refined node representations $\mathrm{X}\!\in\!\mathbb R^{n\times d}$ using the adjusted scores $\beta\!\in\!\mathbb R^{n\times 1}$. The above procedures can be formulated as follows:
\begin{equation}
\label{refine}
    \begin{aligned}
        &\kappa\!=\!\breve{\mathrm{A}}^\ell\mathrm{Z}\mathrm{W}, \quad
        \alpha\!=\!\log(\{\hat{\mathrm{D}}_i\}_{i=1}^n), \\
        &\beta\!=\!\text{softmax}(\delta(\alpha\odot\kappa)), \quad
        \mathrm{X}\!=\!\beta \odot \mathrm{H},
    \end{aligned}
\end{equation}
where $\mathrm{W}\!\in\!\mathbb R^{d\times1}$ is the trainable parameters and $\delta(\cdot)$ is the sigmoid function. $\odot$ denotes the element-wise product.

After completing the node representation learning, we introduce the few-shot setting by defining some key notations. The meta-training tasks $\mathcal{D}_{org}\!=\!\{\mathcal{T}_{t}\}_{t=1}^{\mathrm{T}_{org}}$ are sampled from a task distribution $p(\mathcal{T})$, where each task contains a support set $\mathcal{S}_t\!=\!\{(\mathrm{X}_{t,i}^s,\mathrm{Y}_{t,i}^s)\}_{i=1}^{n_s}$ and a query set $\mathcal{Q}_t\!=\!\{(\mathrm{X}_{t,i}^q,\mathrm{Y}_{t,i}^q)\}_{i=1}^{n_q}$, denoted as $\mathcal{T}_t\!=\!\{\mathcal{S}_t,\mathcal{Q}_t\}$. Here, $\mathrm{X}_{t,i}^\ast$ and $\mathrm{Y}_{t,i}^\ast\!\in\!\mathcal{Y}_{tra}$ denote the node embeddings and its label, where $\mathcal{Y}_{tra}$ denotes the set of base classes. For the meta-testing task $\mathcal{T}_{tes}\!=\!\{\mathcal{S}_{tes},\mathcal{Q}_{tes}\}\!=\!\{\{(\mathrm{X}_{tes,i}^s,\mathrm{Y}_{tes,i}^s)\}_{i=1}^{n_s}, \{(\mathrm{X}_{tes,i}^q,\mathrm{Y}_{tes,i}^q)\}_{i=1}^{n_q}\}$, it is composed in the same way as the meta-training task $\mathcal{T}_t$, with the only difference being that the node label belong to the novel class set $\mathcal{Y}_{tes}$, which is disjoint from $\mathcal{Y}_{tra}$, \textit{i.e.}, $\mathcal{Y}_{tra}\!\cap\!\mathcal{Y}_{tes}\!=\!\emptyset$. When the support set $\mathcal{S}_{tes}$ consists of $N$ sampled classes, each with $K$ nodes, we refer to it as an $N$-way $K$-shot problem. The construction of $\mathcal{Q}_{tes}$ is the same as $\mathcal{S}_{tes}$, except that each class has $M$ nodes. Typically, the model is first trained on the meta-training tasks $\mathcal{D}_{org}$. During the meta-testing stage, the model is fine-tuned on $\mathcal{S}_{tes}$ and then is evaluated the performance on $\mathcal{Q}_{tes}$. 
\subsection{Dual-level Mixup Strategy}
%As stated before, if

%\noindent \textbf{Dual-level Mixup Strategy.}
If we directly conduct few-shot learning on the refined representations, the model's performance would be degraded due to overfitting and constrained generalization. Therefore, we introduce a dual-level mixup strategy, including within-task and across-task mixup, to deal with this issue. Next, we will provide detailed descriptions of each technique.

%\paragraph{Intra-task Mixup.}
%\noindent \textbf{Within-task Mixup.} 
\subsubsection{Within-task Mixup} 
Due to the exceedingly restricted number of sampled nodes in both the support set and query set for each task during the meta-training phase, the efficiency of the meta-training is considerably compromised. Hence, we propose using the within-task mixup strategy to generate more samples for increasing the diversity of the data. Concretely, for a given meta-training task $\mathcal{T}_t$, we perform random sampling on the support set $\mathcal{S}_t$ and query set $\mathcal{Q}_t$, selecting two samples $i$ and $j$ from the same category $k$ for linear interpolation to generate a new one $r$. The above procedure can be formulated as:
\begin{equation}
    \label{intra}
    \begin{aligned}
        \mathrm{X}_{t,r;k}^{\prime s}\!=\!\lambda\mathrm{X}_{t,i;k}^s\!+\! (1\!-\!\lambda)\mathrm{X}_{t,j;k}^s, \quad     \mathrm{X}_{t,r;k}^{\prime q}\!=\!\lambda\mathrm{X}_{t,i;k}^q\!+\! (1\!-\!\lambda)\mathrm{X}_{t,j;k}^q,
    \end{aligned}
\end{equation}
where $\lambda\!\in\![0,1]$ is sampled from the Beta distribution $Beta(\eta,\gamma)$ specified by $\eta$ and $\gamma$.%, \textit{i.e.}, $\lambda\sim Beta(\eta,\gamma)$.

Here, we do not perform label interpolation as the labels of the two sampled nodes are the same, resulting in identical labels for the generated node. There are two reasons for this. First, interpolating samples from different categories would make it difficult to compute prototypes of the mixed labels while expanding the node label space of the original task. Second, this would pose intricate troubles for the subsequent across-task interpolation.

We iteratively apply Eq.\ref{intra} to generate the additional support set $\mathcal{S}_t^\prime\!=\!\{(\mathrm{X}_{t,i}^{\prime s},\mathrm{Y}_{t,i}^s)\}_{i=1}^{n_{s^\prime}}$ and query set $\mathcal{Q}_t^\prime\!=\!\{(\mathrm{X}_{t,i}^{\prime q},\mathrm{Y}_{t,i}^q)\}_{i=1}^{n_{q^\prime}}$, which are subsequently merged with the original corresponding sets to obtain the augmented task $\mathcal{T}_t$ (To avoid introducing extra symbols, we consistently use $\mathcal{T}_t$ to denote the task that undergoes within-task mixup in the following sections.), \textit{i.e.}, $\mathcal{T}_t\!=\!\{\mathcal{S}_t\!\cup\!\mathcal{S}_t^\prime,\mathcal{Q}_t\!\cup\!\mathcal{Q}_t^\prime\}$. The number of nodes in the amplified support and query sets of the augmented task $\mathcal{T}_t$ are $m^\prime\!=\!n_s\!+\!n_{s^\prime}$ and $m\!=\!n_q\!+\!n_{q^\prime}$, respectively.

%\paragraph{Inter-task Mixup.}
%\noindent \textbf{Across-task Mixup.} 
\subsubsection{Across-task Mixup} 
Solely conducting within-task mixup does not address the issue of the limited number of tasks. Therefore, we utilize across-task mixup to directly create new tasks, densifying the task distribution. Specifically, \textit{in the first step}, we randomly select two tasks, $\mathcal{T}_i$ and $\mathcal{T}_j$, from the given meta-training tasks $\mathcal{D}_{org}\!=\!\{\mathcal{T}_t\}_{t=1}^{\mathrm{T}_{org}}$. \textit{In the second step}, we randomly sample a class $k$ from the support set $\mathcal{S}_i$ of $\mathcal{T}_i$ and a class $k^\prime$ from the support set $\mathcal{S}_j$ of $\mathcal{T}_j$, and then compute class-specified support prototypes. This procedure can be expressed as:
%\vspace{-1em}
\begin{equation}
\label{prototype}
    \begin{aligned}
        &\mathrm{C}_{i;k}^s\!=\!\frac{1}{|\mathcal{S}_{i;k}|}\sum_{(\mathrm{X}_{i,\varrho}^s\;,\mathrm{Y}_{i,\varrho}^s)\in\mathcal{S}_{i}}\mathbb{I}_{\mathrm{Y}_{i,\varrho}=k}\mathrm{X}_{i,\varrho}^s \;, \\ &\mathrm{C}_{j;k^\prime}^s\!=\!\frac{1}{|\mathcal{S}_{j;k^\prime}|}\sum_{(\mathrm{X}_{j,\varrho}^s\;,\mathrm{Y}_{j,\varrho}^s)\in\mathcal{S}_{j}}\mathbb{I}_{\mathrm{Y}_{j,\varrho}=k^\prime}\mathrm{X}_{j,\varrho}^s\;,
    \end{aligned}
\end{equation}
where $\mathbb I(\cdot)$ is the indicator function that is 1 when $\mathrm{Y}_{i,\varrho}\!=\!k$, and 0 otherwise. Similarly, we can obtain query prototypes $\mathrm{C}_{i;k}^q$ for class $k$ and $\mathrm{C}_{j;k^\prime}^q$ for class $k^\prime$ by applying Eq.\ref{prototype} to the query sets $\mathcal{Q}_i$ of $\mathcal{T}_i$ and $\mathcal{Q}_j$ of $\mathcal{T}_j$.

\textit{In the third step}, we individually perform feature-level linear interpolation on the support prototypes and query prototypes to generate new samples. Considering that different tasks have different label spaces, we directly treat the label associated with the interpolated data as a new class $\tilde{k}$. We can formulate the above process as:
\begin{equation}
\label{proto_mix}
    \begin{aligned}
        %\tilde{\mathrm{X}}_{t;\tilde{k}}^s\!&=\!\lambda\mathrm{X}_{i;k}^s\!+\!(1\!-\!\lambda)\mathrm{X}_{j;k^\prime}^s, \quad \tilde{\mathrm{Y}}_{t;\tilde{k}}^s\!=\!\Phi(\mathrm{Y}_{i;k}^s,\mathrm{Y}_{j;k^\prime}^s), \\
        %\tilde{\mathrm{X}}_{t;\tilde{k}}^q\!&=\!\lambda\mathrm{X}_{i;k}^q\!+\!(1\!-\!\lambda)\mathrm{X}_{j;k^\prime}^q, \quad \tilde{\mathrm{Y}}_{t;\tilde{k}}^q\!=\!\Phi(\mathrm{Y}_{i;k}^q,\mathrm{Y}_{j;k^\prime}^q), \\
        \tilde{\mathrm{X}}_{t,\varrho;\tilde{k}}^s\!&=\!\lambda\mathrm{C}_{i;k}^s\!+\!(1\!-\!\lambda)\mathrm{C}_{j;k^\prime}^s, \quad \tilde{\mathrm{Y}}_{t,\varrho;\tilde{k}}^s\!=\!\Phi(\mathrm{Y}_{i;k}^s,\mathrm{Y}_{j;k^\prime}^s), \\
        \tilde{\mathrm{X}}_{t,\varrho;\tilde{k}}^q\!&=\!\lambda\mathrm{C}_{i;k}^q\!+\!(1\!-\!\lambda)\mathrm{C}_{j;k^\prime}^q, \quad \tilde{\mathrm{Y}}_{t,\varrho;\tilde{k}}^q\!=\!\Phi(\mathrm{Y}_{i;k}^q,\mathrm{Y}_{j;k^\prime}^q),
    \end{aligned}
\end{equation}
where $\Phi(\cdot,\cdot)$ represents the label uniquely determined by the pair $(\cdot,\cdot)$. We perform $m^\prime$ iterations for the support data and $m$ iterations for the query data in Eq.\ref{proto_mix}, %We execute Eq.\ref{proto_mix} $K$ times to yield $K$-shot samples, 
\textit{i.e.}, $\{\tilde{\mathrm{X}}_{t,\varrho;\tilde{k}}^s,\tilde{\mathrm{Y}}_{t,\varrho;\tilde{k}}^s\}_{\varrho=1}^{m^\prime}$ and $\{\tilde{\mathrm{X}}_{t,\varrho;\tilde{k}}^q,\tilde{\mathrm{Y}}_{t,\varrho;\tilde{k}}^q\}_{\varrho=1}^{m}$. Note that the sampled $\lambda$ each time is different.

\textit{Finally}, we repeat the second and third steps $N$ times to construct an $N$-way $m^\prime$-shot interpolation task $\mathcal{T}_t^{aug}\!=\!\{\tilde{\mathcal{S}}_t,\tilde{\mathcal{Q}}_t\}\!=\!\{\{\tilde{\mathrm{X}}_{t;\tilde{k}}^s,\tilde{\mathrm{Y}}_{t;\tilde{k}}^s\}_{\tilde k=1}^N, \{\tilde{\mathrm{X}}_{t;\tilde{k}}^q, \tilde{\mathrm{Y}}_{t;\tilde{k}}^q\}_{\tilde k=1}^N\}$. We can conduct the above process multiple times to obtain the interpolated tasks $\mathcal{D}_{aug}\!=\!\{\mathcal{T}_t^{aug}\}_{t=1}^{\mathrm{T}_{aug}}$ and merge them with the original tasks $\mathcal{D}_{org}$ to form the final meta-training tasks $\mathcal{D}_{all}\!=\!\mathcal{D}_{org}\!\cup\!\mathcal{D}_{aug}$. The number of tasks in $\mathcal{D}_{all}$ is $\mathrm{T}\!=\!\mathrm{T}_{org}\!+\!\mathrm{T}_{aug}$.
%\subsection{Model Training}

\noindent \textbf{Model Training.}
After performing the dual-level mixup operation, we adopt a classic metric-based episodic training for few-shot node classification. We first derive the prototype $\mathrm{C}_k$ in the support set $\mathcal{S}_t$ of each task $\mathcal{T}_t$ from $\mathcal{D}_{all}$ with the manner shown in Eq.\ref{prototype}.
% \begin{equation}
% \label{compute}
%     \mathrm{C}_k\!=\!\frac{1}{|\mathcal{S}_{t,k}|}\sum_{(\mathrm{X}_{t,i}^s,\mathrm{Y}_{t,i}^s)\in\mathcal{S}_{t}}\mathbb{I}_{\mathrm{Y}_{t,i}=k}\mathrm{X}_{t,i}^s,
% \end{equation}
% where $\mathbb I(\cdot)$ is the indicator function that is 1 when $\mathrm{Y}_{t,i}\!=\!k$, and 0 otherwise.

Next, we optimize the parameters of the model by performing distance-based cross-entropy loss function on all query sets in $\mathcal{D}_{all}$ as:
\begin{equation}
\label{proto}
    \mathcal{L}=%\mathbb E_{\mathcal{T}\sim p(\mathcal{T})}\left[-\sum_{i,k}\log p(\mathrm{Y}_i^q=k|\mathrm{X}_i^q)\right]=\mathbb E_{\mathcal{T}}
\sum_{t=1}^\mathrm{T}\sum_{i=1}^{m}\mathbb I_{\mathrm{Y}_{t,i}=k}\log\frac{\exp(-d(\theta^\top\mathrm{X}_{t,i}^q,\mathrm{C}_k))}{\sum\nolimits_{k^\prime}\exp(-d(\theta^\top\mathrm{X}_{t,i}^q,\mathrm{C}_{k^\prime}))},
\end{equation}
where $d(\cdot,\cdot)$ is the Euclidean distance function and $\theta$ is the trainable vector.

In the meta-testing stage, we do not perform any mixup operations for the evaluated task $\mathcal{T}_{tes}$. Actually, we first use the well-trained model to compute class prototypes on the support set, and then assign samples in the query set to their nearest prototype, defined as: %we fine-tune the well-trained model on the $\mathcal{S}_{tes}$ and apply it into $\mathcal{Q}_{tes}$ to obtain the model performance.
\begin{equation}
    \label{meta-test}
    \begin{aligned}
        &\mathrm{C}_{k}\!=\!\frac{1}{|\mathcal{S}_{tes,k}|}\sum_{(\mathrm{X}_{tes,i}^s,\mathrm{Y}_{tes,i}^s)\in\mathcal{S}_{tes}}\mathbb{I}_{\mathrm{Y}_{tes,i}=k}\mathrm{X}_{tes,i}^s, \\        &\mathrm{Y}_{tes,\ast}^q\!=\!\text{argmin}_kd(\theta^\top\mathrm{X}, \mathrm{C}_k).
    \end{aligned}
\end{equation}

We present the process of proposed SMILE in \textbf{Appendix} \ref{training_procedure}. The time complexity analysis of SMILE are presented in \textbf{Appendix} \ref{complexity}.
\section{Theoretical Analysis}
\label{theorem_section}
In this section, we theoretically analyze why our proposed SMILE, equipped with intra-task and inter-task mixup, can alleviate overfitting and exhibit better generalization capabilities. We first present the obtained key points: \textit{SMILE can regularize the weight parameters in a distribution-dependent manner and reduce the upper bound of the generalization gap by controlling the Rademacher complexity.} Next, we elaborate on the proposed theorems to support the aforementioned points. %For a better analysis of Smile's behavior and for
For simplicity, we conduct a detailed theoretical analysis of SMILE in the binary classification scenario, assuming the use of preprocessed centralized dataset that satisfies the condition $\mathbb E_{\mathrm{X}}[\mathrm{X}]\!=\!0$. Moreover, the proposed SMILE can be modeled as $f_\theta(\mathrm{Z})\!=\!\theta^\top g_\zeta(\mathrm{Z})\!=\!\theta^\top \mathrm{X}$, where $g_\zeta(\cdot)$ denotes the graph encoder parameterized by $\zeta$. We consider using the loss from Eq.\ref{proto} for tasks in $\mathcal{D}_{all}$.
Particularly, the empirical loss function based on enriched training samples for binary classification can be simplified as: 
\begin{equation}
\label{d_loss}
\begin{aligned}
    &\mathcal{L}(\mathcal{D}_{all};\theta)\!=\!\sum_{t=1}^\mathrm{T}\sum_{i=1}^m(1\!+\!\exp(\langle\mathrm{X}_{t,i}^q\!-\!(\mathrm{C}_1\!+\!\mathrm{C}_2)/2,\theta\rangle))^{-1}, \\
    &\mathrm{C}_k\!=\!\frac{1}{|\mathcal{S}_{t,k}|}\sum_{(\mathrm{X}_{t,i}^s,\mathrm{Y}_{t,i}^s)\in\mathcal{S}_{t}}\mathbb{I}_{\mathrm{Y}_{t,i}=k}\mathrm{X}_{t,i}^s,
\end{aligned}
\end{equation}
where $\langle\cdot,\cdot\rangle$ denote the dot product operation. The approximation of the loss function $\mathcal{L}(\mathcal{D}_{all};\theta)$ in Eq.\ref{d_loss} is formalized as:
\begin{equation}
    \label{loss_appro}\mathcal{L}(\mathcal{D}_{all};\theta)\approx
    \mathcal{L}(\mathcal{D}_{org};\theta)\!+\!\mathcal{L}(\bar{\lambda}\mathcal{D}_{org};\theta)\!+\!\mathcal{M}(\theta),
\end{equation}
where $\bar{\lambda}\!=\!\mathbb E_{\rho_\lambda}[\lambda]$ and $\mathcal{M}(\theta)$ is a quadratic regularization term with respect to $\theta$, defined as: 
\begin{equation}
\label{m_theta}
\begin{aligned}
    \mathcal{M}(\theta)\!=\!\mathbb E_{\mathcal{T}_t\sim p(\mathcal{T})}\mathbb E_{(\mathrm{X}_t,\mathrm{Y}_t)\sim q(\mathcal{T}_t)} \frac{\phi(\mathrm{P}_t)(\phi(\mathrm{P}_t)\!-\!0.5)}{2(1\!+\!\exp{(\mathrm{P}_t)})}(\theta^\top\Sigma_\mathrm{X}\theta)
\end{aligned}
\end{equation}
in which $\mathrm
P_t\!=\!\langle\mathrm{X}_t^q \! - \! (\mathrm{C}_1 \! + \! \mathrm{C}_2)/2,\theta\rangle$, $\phi(\mathrm{P}_t)\!=\!\exp(\mathrm{P}_t)/(1\!+\!\exp(\mathrm{P}_t))$, and $\Sigma_\mathrm{X}\!=\!\mathbb E[\mathrm{X}\mathrm{X}^\top]\!=\!\frac{1}{m}\sum\nolimits_{i=1}^m\mathrm{X}_i\mathrm{X}_i^\top$. %The detailed proofs for Eqs.\ref{loss_appro} and \ref{m_theta} can be found in Lemma \ref{lemma_first} of Appendix \ref{proof}.

Eq.\ref{loss_appro} shows that SMILE imposes an additional regularization term on the trainable weights to constrain the solution space, thereby reducing the likelihood of overfitting.

To define the generalization gap problem formally, we introduce a function class of the dual form related to the regularization term in Eq.\ref{loss_appro}, as shown in Eq.\ref{dual}.
\begin{equation}
\label{dual}
\mathcal{F}_\nu\!=\!\{\mathrm{X}\rightarrow\theta^\top\mathrm{X}:\theta^\top\Sigma_\mathrm{X}\theta\leq\nu\}.
\end{equation}
Moreover, we represent the expected risk $\mathsf R$ and empirical risk $\hat{\mathsf R}$ as follows:
\begin{equation}
    \begin{aligned}
        \mathsf R\!=\!\mathbb E_{\mathcal{T}_i \sim p(\mathcal{T})}\mathbb E_{(\mathrm{X}_j,\mathrm{Y}_j)\sim q(\mathcal{T}_i)}\mathcal{L}(f_\theta(\mathrm{X}_j),\mathrm{Y}_j), \\
        \hat{\mathsf R}\!=\!\mathbb E_{\mathcal{T}_i\sim\hat{p}(\mathcal{T})}\mathbb E_{(\mathrm{X}_j,\mathrm{Y}_j)\sim\hat{q}(\mathcal{T}_i)}\mathcal{L}(f_\theta(\mathrm{X}_j),\mathrm{Y}_j).
    \end{aligned}
\end{equation}
Then, we present the following theorem for improved generalization gap brought by SMILE.
\begin{theorem}
\label{theorem_1}
  Assume that $\mathrm{X}, \mathrm{Y}$ and $\theta$ are bounded. For all $f\!\in\! \mathcal{F}_\nu$, where $\theta$ satisfies $\theta^\top\Sigma_\mathrm{X}\theta\!\leq\! \nu$, we have the following generalization gap bound, with probability at least $(1-\epsilon)$ over the training samples,
  %\begin{equation}
  % \begin{align*}
  %   &|\hat{\mathsf{R}}-\mathsf{R}|\leq 2\left(\sqrt{\frac{\nu\cdot\textrm{rank}(\sum_\mathrm{X})}{m}} + \sqrt{\frac{\nu}{\mathrm{T}}}\left(\left\Vert\Sigma_\mathrm{X}^{\dagger/2}\mu_\mathrm{X}\right\Vert + \\
  %   &\textrm{rank}(\Sigma_\mathrm{X}) \right) \right) + 3\left(\sqrt{\frac{(b-a)^2\log(2/\epsilon)}{2m}}+\sqrt{\frac{(c-d)^2\log(2/\epsilon)}{2\mathrm{T}}}\right)
  % \end{align*}
  %\end{equation}
  \begin{equation}
  \begin{split}
    &|\hat{\mathsf{R}}-\mathsf{R}|\leq 2\Bigg(\sqrt{\frac{\nu\cdot\textrm{rank}(\sum_\mathrm{X})}{m}} \!+\! \sqrt{\frac{\nu}{\mathrm{T}}}\bigg( \textrm{rank}(\Sigma_\mathrm{X}) \bigg) \Bigg) \\ 
    &\!+\!  3\left(\sqrt{\frac{\log(2/\epsilon)}{2m}}\!+\!\sqrt{\frac{\log(2/\epsilon)}{2\mathrm{T}}}\right),
  \end{split}
  \end{equation}
  where $m$ and $\mathrm{T}$ denote the number of nodes in the query set and the number of meta-training tasks.
\end{theorem}
%On the one hand, according to Lemma \ref{lemma_first}, the weight space induced by our method are regularized, leading to a smaller μ. On the other hand, the task interpolation and task transfer interpolation increase m and T. These two aspects work together to help reduce the generalization error of the model and alleviate the overfitting problem.

Based on Theorem \ref{theorem_1}, we can obtain several in-depth findings. On the one hand, SMILE induces a regularized weight space for $\theta$, leading to a smaller $\nu$. On the other hand, the introduced intra-task and inter-task interpolations increase $m$ and $\mathrm{T}$ simultaneously. These two aspects work together to reduce the upper bound of the generalization gap of SMILE and alleviate overfitting.

According to the above theorem, we can naturally confirm the following corollary.

\begin{corollary}
\label{corollary_1}
    Let $|\hat{\mathsf R}\!-\!\mathsf R|$ and $|\hat{\mathsf R}_\mathrm{ori}\!-\!\mathsf R_\mathrm{ori}|$ denote the model generalization bounds trained under our proposed task augmentation strategy and standard training strategy, respectively. %obtained from training using our proposed task augmentation strategy and standard training, respectively. 
    We have the following inequality holding,
    \begin{equation}
    \mathsf U (|\hat{\mathsf R}\!-\!\mathsf R|) \leq \mathsf U (|\hat{\mathsf R}_{\mathrm{ori}}\!-\!\mathsf R_{\mathrm{ori}}|),
    \end{equation}
    where $\mathsf U(\cdot)$ denotes the operation of taking the upper bound.
\end{corollary}

Corollary \ref{corollary_1} suggests that SMILE achieves tight generalization bound than other models trained in a standard way. %the upper bound of generalization gap for models trained with our strategy is often smaller than that for models trained with standard training methods.

Suppose the empirical distribution of source tasks in the meta-training be $\mathbb{\hat{P}}$ %, the empirical distribution of the support set for target tasks in the meta-testing be $\mathbb{\hat{Q}}$, 
and the expected distribution of the query set of target tasks be $\mathbb Q$, then during the adaptation process, we expect to reduce the data-dependent upper bound, defined as $\underset{f\in\mathcal{F}}{\text{sup}}|\mathbb E_{\hat{\mathbb P}} \!-\! \mathbb E_{\mathbb Q}|$.
Empirically, when source tasks and target tasks are more similar, the model is more likely to extract generalizable meta-knowledge from source tasks to quickly adapt to target tasks. Theoretically, we present the Theorem \ref{theorem_2}, which demonstrates the reduction of the upper bound between $\mathbb{\hat{P}}$ and $\mathbb{Q}$ induced by our proposed strategy. %We have the following theorem demonstrating the improvement in  induced by our strategy.
\begin{theorem}
\label{theorem_2}
Assume the source tasks and target tasks are drawn from distribution $\mathbb{\hat{P}}$ and $\mathbb{Q}$, and they are independent. For $\epsilon\!>\!0$, with probability at least $(1\!-\!\epsilon)$ over the draws of samples, we have the following upper bound between data distributions,
    \begin{equation}
    \begin{split}
      \underset{f\in\mathcal{F}}{\mathrm{sup}}|\mathbb{E}_{\hat{\mathbb P} }\!-\!\mathbb E_{\mathbb Q}| \leq \Bigg(2\sqrt{\nu\cdot\text{rank}(\Sigma_{\mathrm{X}})} + \sqrt{\frac{\log(1/\epsilon)}{2}} \Bigg) \left( \sqrt{\frac{1}{m}} \!+\! \sqrt{\frac{1}{n_q}} \right),  
    \end{split}
    \end{equation}
    where $n_q$ denotes the number of nodes in the query set of the meta-testing task.
\end{theorem}
We can draw the conclusion that the introduction of intra-task interpolation leads to an increase in the value of $m$. Additionally, according to Eq.\ref{loss_appro}, the regularization effect can result in a decrease of $\nu$. Consequently, Theorem \ref{theorem_2} suggests that our method has the capability to diminish the disparity between the distributions of the source task and target task, facilitating the extraction of pertinent knowledge and, in turn, enhancing the model's generalization. %All detailed proofs can be found in \textbf{Appendix} \ref{proof}.
\section{Experiment}
%\paragraph{Datasets.} 
\noindent \textbf{Datasets.}
%\subsection{Dataset}
To demonstrate the effectiveness of our approach, we conduct few-shot node classification with fewer tasks on four selected prevalent datasets widely used by previous researches, including \textbf{Amazon-Clothing} \cite{mcauley2015inferring}, \textbf{CoraFull} \cite{bojchevski2017deep}, \textbf{Amazon-Electronics} \cite{mcauley2015inferring}, and \textbf{DBLP} \cite{tang2008arnetminer}. Table \ref{dataset} shows the statistics of these datasets. Concisely, \# Nodes and \# Edges represent the number of nodes and edges in the dataset, respectively. \# Features denotes the dimension of the initialized node features, and \# Labels is the number of classes. Class Splits represents the number of classes used for meta-training/meta-validation/meta-testing. 
The detailed descriptions of these evaluated datasets can be found in \textbf{Appendix} \ref{dataset_description}.

\begin{table}[ht]
\centering
\caption{Statistics of the datasets.}
\label{dataset}
\resizebox{0.45\textwidth}{!}{%
\begin{tabular}{@{}c|ccccc@{}}
\toprule
Dataset            & \# Nodes & \# Edges & \# Features & \# Labels & Class Splits \\ \midrule
Amazon-Clothing    & 24,919   & 91,680   & 9,034       & 77        & 40/17/20     \\
Cora-Full          & 19,793   & 65,311   & 8,710       & 70        & 25/20/25     \\
Amazon-Electronics & 42,318   & 43,556   & 8,669       & 167       & 90/37/40     \\
DBLP               & 40,672   & 288,270  & 7,202       & 137       & 80/27/30     \\ \bottomrule
\end{tabular}%
}
\end{table}

%\paragraph{Baselines.} 
\noindent \textbf{Baselines.} 
%\subsection{Baselines}
We mainly select three types of baselines for comparison to verify the superiority of the proposed SMILE. \textit{Traditional meta-learning} methods consist of \textbf{Protonet} \cite{snell2017prototypical} and \textbf{MAML} \cite{finn2017model}. \textit{Meta-learning with fewer tasks} methods comprise \textbf{MetaMix} \cite{yao2021improving}, \textbf{MLTI} \cite{yao2021meta}, and \textbf{Meta-Inter} \cite{lee2022set}. \textit{Graph meta-learning} methods include \textbf{Meta-GNN} \cite{zhou2019meta}, \textbf{GPN} \cite{ding2020graph}, \textbf{G-Meta} \cite{huang2020graph}, \textbf{Meta-GPS} \cite{liu2022few}, \textbf{X-FNC} \cite{wang2023few}, \textbf{COSMIC} \cite{wang2023contrastive}, \textbf{TLP} \cite{tan2022transductive}, and \textbf{TEG} \cite{kim2023task}. %\textit{Graph prompting} methods contain \textbf{GraphPrompt} \cite{liu2023graphprompt}, and \textbf{GPF} \cite{fang2024universal}. 
The descriptions and implementations of these baselines are provided in \textbf{Appendix} \ref{baseline}. 

%\paragraph{Implementation Details.} 
\noindent \textbf{Implementation Details.} In the \textit{node representation learning} stage, we adopt 2-layer SGC with 16 hidden units. In the \textit{dual-level mixup} stage, we uniformly set the two parameters involved in the beta distribution to 0.5, \textit{i.e.}, $\eta\!=\!\gamma\!=\!0.5$. Moreover, in within-task mixup, we generate the same number of nodes as the original support and query set in each meta-training task by default, that is, $n_{s^\prime}\!=\!n_s$, $n_{q^\prime}=n_q$. In across-task mixup, we generate as many interpolated tasks as original tasks, that is $\mathrm{T}_{aug}\!=\!\mathrm{T}_{org}$. %For the graph meta-learning baselines, we use the hyperparameters recommended in the original papers. For traditional meta-learning models, we conduct careful hyperparameter search and report their optimal performance. Particularly, 
In the cross-domain setting, we meta-train the model on one source domain and then meta-test it on another target domain. To ensure fair comparison, we perform meta-training on the same sampled tasks for all models. %For those meta-learning with fewer tasks baselines, we uniformly use SGC as the graph encoder. 
Moreover, we evaluate the performance of our model using the average accuracy (Acc) and macro-F1 (F1) score across 50 randomly selected meta-testing tasks.
\section{Result}
\begin{table*}[ht]
\small
\centering
\caption{Results (\%) of different models using fewer tasks on datasets under the 5-way 5-shot \textit{in-domain} setting. Bold: best (based on the pairwise t-test with 95\% confidence). Underline: runner-up.} %Rel. Imp.: relative improvement over the second-best model.}
\label{res_one}
\resizebox{0.9\textwidth}{!}{%
\begin{tabular}{@{}c|cccccccc|cccccccc@{}}
\toprule
\multirow{3}{*}{Model} & \multicolumn{8}{c|}{Amazon-Clothing}                                                                                                  & \multicolumn{8}{c}{CoraFull}                                                                                                         \\ \cmidrule(l){2-17} 
                       & \multicolumn{2}{c}{5 tasks}     & \multicolumn{2}{c}{10 tasks}    & \multicolumn{2}{c}{15 tasks}    & \multicolumn{2}{c|}{20 tasks}   & \multicolumn{2}{c}{5 tasks}     & \multicolumn{2}{c}{10 tasks}    & \multicolumn{2}{c}{15 tasks}    & \multicolumn{2}{c}{20 tasks}    \\ \cmidrule(l){2-17} 
                       & Acc            & F1             & Acc            & F1             & Acc            & F1             & Acc            & F1             & Acc            & F1             & Acc            & F1             & Acc            & F1             & Acc            & F1             \\ \midrule
Protonet               & 49.17          & 48.36          & 53.51          & 52.55          & 55.82          & 54.99          & 57.99          & 57.14          & 37.20          & 35.98          & 40.14          & 38.89          & 43.90          & 42.96          & 45.58          & 44.34          \\
MAML                   & 44.90          & 43.66          & 45.67          & 44.44          & 46.29          & 44.97          & 46.90          & 45.60          & 38.15          & 36.83          & 42.26          & 41.28          & 44.21          & 43.95          & 46.37          & 45.43          \\ \midrule
MetaMix                   & 78.32          & 78.22          & 78.66          & 78.52           & 80.16          & 79.15          & 81.09          & 80.52          & 62.95          & 62.25          & 64.20          & 63.95          & 65.72          & 64.19          & 67.59          & 66.26          \\
MLTI                   & 79.19          & 78.59          & 79.91          & 78.92           & 80.22          & 79.39          & 81.27          & 80.86          & 63.19          & 63.06          & 65.72          & 65.69          & 66.25          & 64.92          & 67.15          & 66.10          \\
Meta-Inter                   & \underline{79.92}          & \underline{79.22}          & \underline{80.12}          & \underline{79.56}           & \underline{80.55}          & \underline{79.90}          & \underline{81.26}          & \underline{81.05}          & \underline{63.82}          & \underline{63.36}          & 66.59          & 65.92          & 67.19          & 65.50          & 68.22          & 67.59          \\ \midrule
Meta-GNN               & 55.29          & 50.44          & 57.19          & 53.65          & 62.29          & 59.55          & 70.19          & 67.22          & 42.96          & 40.83          & 45.09          & 42.87          & 47.15          & 45.38          & 49.88          & 48.12          \\
GPN                    & 68.23          & 67.16          & 70.06          & 69.57          & 72.40          & 71.95          & 72.81          & 71.56          & 43.35          & 42.08          & 46.19          & 44.81          & 51.56          & 50.24          & 55.83          & 54.76          \\
G-Meta                 & 60.43          & 60.11          & 64.51          & 63.74          & 68.99          & 67.96          & 71.98          & 72.75          & 45.84          & 44.27          & 49.22          & 48.91          & 51.15          & 50.53          & 59.12          & 58.56          \\
Meta-GPS               & 62.02          & 59.76          & 69.21          & 69.04          & 73.01          & 71.92          & 75.74          & 74.85          & 50.33          & 48.22          & 57.85          & 54.86          & 61.28          & 60.11          & 63.76          & 62.28          \\
%TENT                   & 73.15    & 70.99    & 75.77    & 72.19    & 79.90    & 78.73    & 81.90    & 79.36    &52.62    &50.15    &59.69    &56.69    & 66.88    & 64.44    & 68.90    & 66.46    \\ 
X-FNC                   & 69.12    & 68.29    & 72.12    & 71.11    & 75.19    & 74.63    & 79.26    & 78.02    & 55.06          & 53.10          & 61.53          & 60.29          & 65.22    & 64.10    & 66.09    & 65.12    \\ 
COSMIC                   & 75.66    & 74.92    & 76.39    & 75.72    & 77.92    & 76.59    & 78.36    & 77.39    & 62.29          & 60.39          & 65.39          & 64.80          & 66.72    & 65.72    & 68.29    & 67.20
\\
TLP                   & 71.39    & 70.39    & 73.39    & 72.52    & 74.72    & 73.36    & 75.60    & 74.29    & 51.79          & 49.72          & 56.72          & 55.79          & 57.72    & 56.73    & 57.99    & 57.30
\\ 
TEG                   & 78.55    & 77.92    & 80.26    & 79.30    & 80.82    & 79.99    & 81.19    & 80.16    & 62.89          & 61.26          & \underline{68.29}          & \underline{67.39}          & \underline{68.59}    & \underline{67.55}    & \underline{70.06}    & \underline{69.29}
\\ \midrule
%GraphPrompt                   & 70.92    & 69.72    & 72.39    & 71.99    & 73.92    & 73.26    & 74.28    & 73.55    & 50.12          & 49.72          & 53.72          & 52.19          & 56.72    & 55.18    & 57.72    & 56.19 \\
%GPF                    & 71.96    & 70.39    & 72.23    & 71.56    & 74.72    & 73.92    & 75.38    & 74.39    & 51.39          & 50.09          & 54.19          & 53.72          & 56.30    & 55.92    & 57.92    & 56.92 \\ \midrule
SMILE                   & \textbf{82.80} & \textbf{82.49} & \textbf{83.46} & \textbf{82.88} & \textbf{83.92} & \textbf{83.33} & \textbf{84.66} & \textbf{84.52} & \textbf{66.34} & \textbf{65.70} & \textbf{71.72} & \textbf{71.15} & \textbf{70.78} & \textbf{70.19} & \textbf{72.60} & \textbf{72.10} \\ \midrule
%Rel. Imp.              & 13.19          & 16.20          & 10.15           & 14.81          & 5.03           & 5.84           & 3.37           & 4.60           & 20.49          & 23.73          & 16.56          & 18.01          & 5.83           & 8.92           & 5.37           & 8.49           \\ \midrule
\multirow{3}{*}{Model} & \multicolumn{8}{c|}{Amazon-Electronics}                                                                                               & \multicolumn{8}{c}{DBLP}                                                                                                              \\ \cmidrule(l){2-17} 
                       & \multicolumn{2}{c}{5 tasks}     & \multicolumn{2}{c}{10 tasks}    & \multicolumn{2}{c}{15 tasks}    & \multicolumn{2}{c|}{20 tasks}   & \multicolumn{2}{c}{5 tasks}     & \multicolumn{2}{c}{10 tasks}    & \multicolumn{2}{c}{15 tasks}    & \multicolumn{2}{c}{20 tasks}    \\ \cmidrule(l){2-17} 
                       & Acc            & F1             & Acc            & F1             & Acc            & F1             & Acc            & F1             & Acc            & F1             & Acc            & F1             & Acc            & F1             & Acc            & F1             \\ \midrule
Protonet               & 46.20          & 45.09          & 49.56          & 48.57          & 51.98          & 51.05          & 54.03          & 53.20          & 46.57          & 45.47          & 50.90          & 49.81          & 51.02          & 49.74          & 52.09          & 51.05          \\
MAML                   & 34.34          & 33.42          & 34.76          & 33.76          & 35.42          & 34.41          & 35.91          & 34.95          & 39.71          & 38.86          & 40.34          & 39.58          & 40.70           & 39.85          & 41.31          & 40.58          \\ \midrule
MetaMix                   & 61.96          & 61.82          & 63.72          & 63.66          & 65.19          & 64.92          & 66.15          & 65.72          & 72.12          & 71.15          & 73.19          & 72.12          & 75.16          & 73.95          & 76.22          & 74.79          \\
MLTI                   & 62.25          & 62.02          & 65.26          & 65.09          & 66.72          & 65.59          & 67.19          & 66.22          & 72.36          & 71.96          & 72.92          & 72.55          & 73.22          & 73.10          & 75.10          & 74.95          \\
Meta-Inter                   & 62.79          & 62.56          & 65.76          & 65.52          & 67.19          & 66.15          & 68.99          & 67.29          & 72.52          & 72.11          & 73.19          & 72.99          & 74.28          & 73.25          & 75.29          & 75.10          \\ \midrule
Meta-GNN               & 40.52          & 39.74          & 46.16          & 45.87          & 48.92          & 47.93          & 50.86          & 50.07          & 50.68          & 49.04          & 53.86          & 49.67          & 59.72          & 59.36          & 65.49          & 62.12          \\
GPN                    & 49.08          & 47.91          & 51.12          & 49.98          & 54.24          & 53.23          & 56.69          & 55.62          & 70.26    & 69.13    & \underline{74.42}    & \underline{73.48}    & \underline{76.02}          & \underline{75.03}          & \underline{76.61}          & \underline{75.60}          \\
G-Meta                 & 43.29          & 42.20          & 49.57          & 52.90          & 56.96          & 55.38          & 60.41          & 59.91          & 53.08          & 48.13          & 55.92          & 53.64          & 57.82          & 56.76          & 63.17          & 62.85          \\
Meta-GPS               & 46.11          & 43.62          & 57.90          & 56.20          & 67.73          & 66.69          & 70.13          & 69.15          & 56.59          & 54.12          & 65.20          & 63.20          & 73.00          & 72.35          & 75.16          & 73.19          \\
%TENT                   & \underline{63.36}    & \underline{63.59}    & \underline{69.70}    & \underline{67.12}    & \underline{70.09}    & \underline{68.60}    & \underline{72.09}    & \underline{71.60}    & 70.15          & 69.02          & 74.31          & 72.22          & \underline{76.89}    & \underline{76.72}    & \underline{77.87}    & \underline{77.70}    \\ 
X-FNC                   & 59.26    & 56.39    & 63.72    & 62.10    & \underline{69.82}    & 67.63    & 71.36    & 70.02    & 69.06          & 68.10          & 72.53          & 71.29          & 74.29    & 73.22    & 76.19    & 75.20    \\ 
COSMIC                   & 64.06    & 63.02    & \underline{67.36}    & \underline{66.32}    & 68.22    & 67.09    & 70.16    & 69.30    & 71.29          & 70.19          & 72.09          & 70.80          & 73.02    & 71.20    & 75.16    & 72.22    \\
TLP                   & 63.09    & 62.19    & 64.30    & 63.59    & 65.72    & 64.32    & 67.18    & 66.72    & 71.26          & 70.75          & 72.87          & 72.09          & 73.39    & 73.06    & 75.16    & 74.69    \\
TEG                   & \underline{65.90}    & \underline{64.62}    & 67.29    & 66.22    & 69.80    & \underline{68.29}    & \underline{72.12}    & \underline{71.16}    & \underline{72.59}          & \underline{72.26}          & 73.79          & 72.19          & 75.52    & 74.50    & 76.26    & 75.12    \\
\midrule
%GraphPrompt                   & 59.90    & 57.92    & 60.19    & 59.29    & 62.02    & 61.25    & 64.18    & 63.25    & 66.92          & 65.72          & 67.70          & 66.12          & 69.72    & 68.28    & 70.27    & 69.59 \\
%GPF                    & 59.96    & 58.99    & 61.93    & 60.96    & 62.12    & 61.96    & 63.82    & 63.29    & 66.52          & 65.59          & 70.12          & 69.92          & 70.29    & 70.29    & 71.12    & 70.92 \\ \midrule
SMILE                   & \textbf{67.30} & \textbf{66.30} & \textbf{70.76} & \textbf{70.05} & \textbf{73.48} & \textbf{72.66} & \textbf{75.42} & \textbf{75.42} & \textbf{75.88} & \textbf{75.05} & \textbf{76.64} & \textbf{75.77} & \textbf{79.56} & \textbf{78.77} & \textbf{80.50} & \textbf{79.61} \\ \bottomrule
%Rel. Imp.              & 6.22           & 4.62           & 1.52           & 4.37           & 4.84           & 5.92           & 4.62           & 5.34           & 8.00           & 8.56           & 2.98           & 3.12           & 3.47           & 2.67           & 3.38           & 2.46           \\ \bottomrule
\end{tabular} %
}
\end{table*}

\begin{table*}[ht]
\centering
\small
\caption{Results (\%) of different models using fewer tasks on datasets under the 5-way 5-shot \textit{cross-domain} setting. A$\rightarrow$B denotes the model is trained on A and evaluated on B.}
\label{res_cross}
\resizebox{0.9\textwidth}{!}{ %
\begin{tabular}{@{}c|cccccccc|cccccccc@{}}
\toprule
\multirow{3}{*}{Dataset} & \multicolumn{8}{c|}{Amazon-Clothing$\rightarrow$CoraFull}                                                                           & \multicolumn{8}{c}{CoraFull$\rightarrow$Amazon-Clothing}                                                                            \\ \cmidrule(l){2-17} 
                         & \multicolumn{2}{c}{5 tasks}     & \multicolumn{2}{c}{10 tasks}    & \multicolumn{2}{c}{15 tasks}    & \multicolumn{2}{c|}{20 tasks}   & \multicolumn{2}{c}{5 tasks}     & \multicolumn{2}{c}{10 tasks}    & \multicolumn{2}{c}{15 tasks}    & \multicolumn{2}{c}{20 tasks}    \\ \cmidrule(l){2-17} 
                         & Acc            & F1             & Acc            & F1             & Acc            & F1             & Acc            & F1             & Acc            & F1             & Acc            & F1             & Acc            & F1             & Acc            & F1             \\ \midrule
Protonet                 & 20.72          & 7.90           & 22.84          & 10.89          & 29.70          & 15.91          & 32.96          & 18.19          & 24.84          & 13.18          & 29.96          & 22.49          & 32.84          & 26.01          & 34.58          & 29.90          \\
MAML                     & 20.40          & 13.51          & 20.74          & 13.19          & 26.68          & 12.22          & 30.19          & 15.92          & 23.56          & 12.10          & 27.35          & 20.16          & 30.19          & 23.95          & 32.96          & 27.96          \\ \midrule
MetaMix                   & 31.96          & 28.76          & 33.12          & 31.22          & 35.22          & 33.19          & 37.15          & 35.55          & 34.76          & 31.66          & 36.25          & 33.69          & 39.72          & 37.16          & 41.26          & 39.25 \\
MLTI                   & 33.29          & 30.12          & 35.16          & 32.29          & 38.25          & 35.52          & 40.22          & 38.29          & 35.12          & 33.49          & 37.22          & 35.29          & 42.19          & 41.52          & 45.66          & 43.95          \\
Meta-Inter                   & 34.72          & 32.19         & 35.76          & 34.26          & 40.16          & 37.59          & 42.29          & 40.32          & 41.76          & 39.59          & 43.22          & 41.57          & 44.11          & 42.25          & 47.29          & 45.55          \\ \midrule
Meta-GNN                 & 26.36          & 20.99          & 30.50          & 26.72          & 33.22          & 30.15          & 35.99          & 32.16          & 32.16          & 22.39          & 35.22          & 26.62          & 38.16          & 29.35          & 39.66          & 32.90          \\
GPN                      & 35.86    & 34.81    & 39.38    & 38.03    & 41.10    & 39.82    & 41.96    & 41.15    & 40.08          & 38.73          & 41.78          & 40.67          & 43.90          & 42.87          & 45.04          & 44.30          \\
G-Meta                   & 30.36          & 26.95          & 33.19          & 29.62          & 35.29          & 33.16          & 36.21          & 35.20          & 35.22          & 30.16          & 37.22          & 30.29          & 40.19          & 32.29          & 41.19          & 36.96          \\
Meta-GPS                 & 32.02          & 27.07          & 34.15          & 30.19          & 35.66          & 34.15          & 39.26          & 37.55          & 45.59          & 43.29          & 47.62          & 45.10          & 50.19          & 47.12          & 52.19          & 49.32          \\
%TENT                     & 34.37          & 32.75          & 36.17          & 33.71          & 38.68          & 36.08          & 40.56          & 37.91          & \underline{49.29}    & \underline{46.93}    & \underline{52.80}    & \underline{49.72}    & \underline{54.91}    & \underline{52.38}    & \underline{55.69}    & \underline{54.62}    \\
X-FNC                    & 33.59          & 31.10          & 35.15          & 32.19          & 37.25          & 34.12          & 39.72          & 36.29          & 47.26          & 45.16          & 49.30          & 46.22          & 52.20          & 49.29          & 53.72          & 50.22          \\ 
COSMIC                    & \underline{38.02}          & 36.22          & 40.09          & 37.05          & \underline{42.20}          & 39.09          & \underline{42.46}          & 40.30          & 49.20          & 47.19          & 52.02          & 51.29          & 53.09          & 52.16          & \underline{55.39}          & \underline{53.90}          \\
TLP                    & 37.99          & \underline{37.29}          & \underline{41.23}          & \underline{39.59}          & 41.99          & \underline{40.92}          & 42.26          & \underline{41.25}          & \underline{51.12}          & \underline{50.15}          & \underline{53.90}          & \underline{52.29}          & \underline{54.26}          & \underline{52.66}          & 55.20          & 53.30          \\
TEG                    & 33.05          & 31.29          & 35.26          & 34.32          & 35.80          & 34.69          & 36.35          & 35.36          & 41.09          & 40.20          & 42.12          & 41.39          & 43.72          & 42.60          & 46.56          & 43.87          \\
\midrule
%GraphPrompt                    & 32.10          & 31.92          & 35.29          & 34.11          & 36.22          & 35.55          & 37.28          & 36.22          & 42.12          & 40.12          & 44.36          & 42.29          & 47.39          & 45.32          & 49.17          & 48.29          \\
%GPF                    & 32.26          & 31.72          & 34.23          & 33.06          & 35.19          & 35.06          & 37.26          & 36.99          & 42.92          & 41.50          & 43.17          & 42.72          & 45.12          & 44.29          & 46.22          & 45.90          \\ \midrule
SMILE                    & \textbf{42.64} & \textbf{41.27} & \textbf{45.14} & \textbf{43.69} & \textbf{45.88} & \textbf{44.10} & \textbf{46.72} & \textbf{45.65} & \textbf{56.36} & \textbf{55.25} & \textbf{58.84} & \textbf{57.53} & \textbf{59.08} & \textbf{57.96} & \textbf{59.38} & \textbf{58.25} \\ \midrule
%Rel. Imp.                 & 18.91          & 18.56          & 14.62          & 14.88          & 11.63          & 10.75          & 11.34          & 10.93          & 14.34          & 17.73          & 11.44          & 15.71          & 7.59           & 6.83           & 6.63           & 6.65           \\ \midrule
\multirow{3}{*}{Dataset} & \multicolumn{8}{c|}{Amazon-Electronics$\rightarrow$DBLP}                                                                            & \multicolumn{8}{c}{DBLP$\rightarrow$Amazon-Electronics}                                                                             \\ \cmidrule(l){2-17} 
                         & \multicolumn{2}{c}{5 tasks}     & \multicolumn{2}{c}{10 tasks}    & \multicolumn{2}{c}{15 tasks}    & \multicolumn{2}{c|}{20 tasks}   & \multicolumn{2}{c}{5 tasks}     & \multicolumn{2}{c}{10 tasks}    & \multicolumn{2}{c}{15 tasks}    & \multicolumn{2}{c}{20 tasks}    \\ \cmidrule(l){2-17} 
                         & Acc            & F1             & Acc            & F1             & Acc            & F1             & Acc            & F1             & Acc            & F1             & Acc            & F1             & Acc            & F1             & Acc            & F1             \\ \midrule
Protonet                 & 31.86          & 22.56          & 32.58          & 23.73          & 35.90          & 32.76          & 39.88          & 35.71          & 28.84          & 18.62          & 30.54          & 20.08          & 33.10          & 22.37          & 35.46          & 25.20          \\
MAML                     & 29.17          & 19.13          & 30.10          & 22.15          & 32.97          & 25.98          & 35.25          & 29.11          & 26.59          & 17.99          & 28.36          & 19.29          & 30.02          & 20.15          & 32.16          & 22.16          \\ \midrule
MetaMix                   & 40.16          & 35.68           & 43.25          & 41.69          & 45.19          & 43.12          & 49.12          & 43.59          & 37.70          & 35.22          & 40.20          & 39.09          & 42.25         & 40.16          & 44.19          & 42.20          \\
MLTI                   & 42.12          & 37.19          & 46.39          & 45.06          & 49.19          & 47.25          & 51.35          & 50.39          & 38.22          & 36.96          & 41.39          & 40.25          & 43.39          & 42.05          & 46.12          & 45.09          \\
Meta-Inter                   & 46.19          & 45.12          & 48.15          & 46.79          & 51.29          & 49.76          & 53.18          & 51.02          & 41.50          & 40.15          & 43.19          & 41.10          & 45.20          & 43.35          & 47.15          & 45.05          \\ \midrule
Meta-GNN                 & 39.19          & 34.72          & 42.26          & 39.16          & 43.96          & 39.55          & 45.66          & 42.19          & 35.72          & 33.20          & 39.59          & 38.62          & 40.39          & 39.29          & 41.26          & 40.22          \\
GPN                      & \underline{60.08}    & \underline{58.75}    & \underline{61.92}    & \underline{61.58}    & \underline{63.19}    & \underline{62.60}    & \underline{63.99}    & \underline{63.10}    & 42.99          & 41.46          & \underline{46.36}    & \underline{44.73}    & \underline{47.09}    & \underline{45.42}    & \underline{47.52}          & \underline{45.76}          \\
G-Meta                   & 45.72          & 43.32          & 47.22          & 45.09          & 47.96          & 45.99          & 49.56          & 48.39          & 37.22          & 35.93          & 40.19          & 39.52          & 42.35          & 40.19          & 43.62          & 42.19          \\
Meta-GPS                 & 47.59          & 46.70          & 49.20          & 47.16          & 50.26          & 49.96          & 52.39          & 51.22          & \underline{43.06}          & \underline{42.05}          & 45.12          & 43.16          & 46.02          & 44.95          & 46.79          & 45.02          \\
%TENT                     & 50.79          & 49.41          & 51.01          & 49.54          & 53.45          & 51.73          & 54.67          & 53.19          & \underline{43.37}    & \underline{42.94}    & 44.75          & 43.38          & 46.71          & 45.37          & \underline{47.99}    & \underline{46.79}    \\
X-FNC                    & 49.19          & 48.36          & 49.55          & 48.02          & 51.35          & 50.26          & 52.90          & 51.39          & 41.59          & 40.02          & 42.36          & 42.19          & 44.16          & 43.16          & 46.39          & 45.25          \\ 
COSMIC                    & 57.22          & 55.30          & 58.29          & 57.35          & 60.20          & 61.19          & 61.36          & 62.30          & 39.20          & 37.11          & 41.02          & 40.22          & 43.03          & 42.22          & 44.32          & 43.26          \\
TLP                    & 58.25          & 57.19          & 59.33          & 59.01          & 61.29          & 60.02          & 62.16          & 61.25          & 41.11          & 40.16          & 43.20          & 42.25          & 44.16          & 42.32          & 45.20          & 43.90          \\
TEG                    & 38.05          & 36.29          & 40.21          & 39.32          & 42.80          & 41.69          & 43.35          & 42.36          & 33.19          & 31.20          & 34.22          & 33.52          & 35.70          & 34.62          & 36.55          & 35.37          \\ \midrule
%GraphPrompt                    & 55.30          & 53.90          & 56.19          & 55.39          & 57.62          & 56.15          & 59.28          & 56.22          & 37.22          & 36.29          & 38.56          & 37.56          & 40.19          & 39.32          & 41.17          & 40.29          \\
%GPF                    & 54.29          & 53.79          & 56.13          & 55.26          & 57.29          & 55.36          & 59.16          & 56.39          & 38.92          & 37.52          & 41.27          & 40.17          & 41.92          & 40.29          & 42.19          & 40.69          \\ \midrule
SMILE                    & \textbf{62.44} & \textbf{61.66} & \textbf{64.54} & \textbf{64.16} & \textbf{65.04} & \textbf{64.43} & \textbf{65.78} & \textbf{65.42} & \textbf{46.24} & \textbf{44.54} & \textbf{48.82} & \textbf{47.26} & \textbf{49.26} & \textbf{47.70} & \textbf{49.52} & \textbf{47.88} \\ \midrule
%Rel. Imp.                 & 3.93           & 4.95           & 4.23           & 4.19           & 2.93           & 2.92           & 2.80           & 3.68           & 6.62           & 3.73           & 5.31           & 5.66           & 4.61           & 4.79           & 3.19           & 2.33           \\ \bottomrule
\end{tabular} %
}
\end{table*}

%\paragraph{Model Performance.}
\noindent \textbf{Model Performance.} 
%\subsection{Model Performance}
We present the results of our proposed SMILE and other models under both \textit{in-domain} and \textit{cross-domain} settings with different number of tasks across several datasets in Tables \ref{res_one} and \ref{res_cross}. %For readability, we have placed the results with \textit{standard deviations} in \textbf{Appendix} \ref{full_results}.
According to above results, we can obtain the following in-depth analysis. %We present the results of our proposed SMILE and other baselines under 5-way 5-shot experimental setting with fewer tasks across several datasets in Table \ref{res_one}. Also, we show the results of these models under 10-way 5-shot and sufficient meta-training tasks experimental settings in the Appendix \ref{more_experiment}. According to Table \ref{res_one}, we can obtain the following in-depth analysis.
We can find that our approach achieves the best performance across varying numbers of meta-training tasks in both in-domain and cross-domain settings for all datasets, demonstrating its superiority in dealing with graph few-shot learning with fewer tasks. %This phenomenon is in accordance with our conjecture. 
One plausible reason is that we explicitly introduce degree-based prior in the node representation stage, resulting in more discriminative features beneficial for subsequent tasks. Furthermore, we employ a dual-level mixup strategy, enriching the diversity of both within-task and across-task data, effectively alleviating the negative impact of data and task scarcity. These strategies facilitate the model to extract more transferable meta-knowledge, thereby greatly enhancing its generalization capability.

%We find that as the number of available meta-training tasks decreases, the gains of SMILE become increasingly considerable, while the remaining models exhibit an intolerable performance decline. This aligns with our expectations. Taking the 5 tasks setting on the Cora-Full dataset as an example, as shown in Table \ref{res_one}, SMILE outperforms the second-best model X-FNC by a large margin, reaching 20\%. With the scenarios with fewer tasks, our method can densify the task distribution by explicitly increasing the number of meta-training tasks. However, if there are already adequate tasks, it is not difficult for these models to extract transferable knowledge during meta-training to adapt to meta-testing tasks. Thus, the improvements brought by our method decreases. %However, as the number of original tasks increases, the resulting improvement diminishes. Generally, graph meta-learning methods consistently surpass traditional meta-learning methods because the former simultaneously consider both the feature and structure information of the graph.

We find that graph meta-learning models represented by COSMIC and TEG perform well in scenarios with more tasks across multiple datasets, which aligns with our expectations. These models are specifically designed for graph few-shot learning and utilize unique few-shot algorithms that enable them to achieve discriminative node representations with limited labeled data. However, they struggle in scenarios with fewer tasks, performing significantly worse than our model. This is because they can only extract sufficient transferable meta-knowledge when there are ample meta-training tasks. 
Moreover, the meta-learning with fewer tasks models equipped with SGC, such as MLTI and Meta-Inter, also demonstrate impressive performance in the both in-domain and cross-domain experimental settings. We attribute this phenomenon to the specific strategies employed by these models to mitigate the negative effects caused by limited tasks. However, this type of models still significantly lag behind our model, as they do not incorporate degree-based prior knowledge and fail to address scarcity issues from both data and task perspectives simultaneously. Additionally, traditional meta-learning methods consistently underperform compared to other methods because they completely overlook the important structural information in the graph.

%Also, we show additional results of these models under different experimental settings in the \textbf{Appendix} \ref{more_experiment}, including model performance with sufficient tasks and across various graph encoders and so on, due to the space constraints.

%\paragraph{Ablation Study.} 
\noindent \textbf{Ablation Study.} 
%\subsection{Ablation Study}
To demonstrate the effectiveness of our adopted strategies, we design several model variants. (I) \textit{vanilla mixup}: Without the dual-level mixup, we first compute the prototypes using the support set, and then perform vanilla mixup to query set, which involves mixing the features and generates soft labels. (II) \textit{internal mixup}: In the absence of dual-level mixup, we directly perform mixup on different classes within each task and generate corresponding hard labels, treating them as novel classes. %and the generated labels are treated as novel classes. 
(III) \textit{w/o within and across}: We simultaneously discard both within-task and across-task mixup operations. (IV) \textit{w/o within}: We delete the within-task operation and leave the across-task one. (V) \textit{w/o across}: We remove the across-task strategy and retain the within-task one. (VI) \textit{w/o degree}: We exclude the utilization of degree information and solely employ the vanilla SGC for node representation learning.

%Through the analysis of the results presented in 
According to Table \ref{ablation}, it is evident that the employed strategies have a favorable impact on the model performance. The introduction of dual-level mixup enriches the samples within each task and provides diverse tasks, making significant contributions to enhancing the model. The adopted degree-based prior information also improves the model by learning expressive node embeddings, especially in cross-domain setting. When removing the degree information, the performance drastically declines. One plausible reason is that this module explicitly utilizes structural prior knowledge from the target graph domain, benefiting downstream tasks. Additionally, the results of model variants I and II demonstrate that performing label interpolation within each task, whether generating soft or hard labels, degrades the model performance. %The possible reason is that the introduced mixing labels can confuse the model.

\begin{table*}[ht]
\caption{Results of different model variants with respect to 5 tasks under the 5-way 5-shot setting.}
\label{ablation}
\centering
\normalsize
\resizebox{0.9\textwidth}{!}{%
\begin{tabular}{@{}c|cccccccc|cccccccc@{}}
\toprule
\multirow{2}{*}{Dataset} & \multicolumn{2}{c}{Clothing} & \multicolumn{2}{c}{Electronics} & \multicolumn{2}{c}{DBLP}        & \multicolumn{2}{c|}{CoraFull}   & \multicolumn{2}{c}{Clothing$\rightarrow$CoraFull} & \multicolumn{2}{c}{CoraFull$\rightarrow$Clothing} & \multicolumn{2}{c}{Electronics$\rightarrow$DBLP} & \multicolumn{2}{c}{DBLP$\rightarrow$Electronics} \\ \cmidrule(l){2-17} 
                         & Acc              & F1               & Acc                & F1                & Acc            & F1             & Acc            & F1             & Acc                      & F1                       & Acc                      & F1                       & Acc                      & F1                      & Acc                      & F1                      \\ \midrule
vanilla mixup            & 78.82            & 78.45            & 60.98              & 60.70             & 74.06          & 73.03          & 61.90          & 61.07          & 38.90                    & 38.26                    & 52.93                    & 52.29                    & 57.92                    & 57.30                   & 40.52                    & 39.26                   \\
internal mixup           & 78.96            & 78.72            & 63.59              & 62.21             & 75.34          & 74.45          & 63.64          & 63.00          & 39.12                    & 38.59                    & 53.22                    & 53.19                    & 59.02                    & 57.96                   & 41.36                    & 40.56                   \\
w/o within and across    & 79.10            & 78.97            & 62.36              & 61.08             & 73.14          & 72.36          & 62.44          & 61.85          & 39.72                    & 39.19                    & 53.58                    & 52.42                    & 59.34                    & 58.94                   & 42.12                    & 41.41                   \\
w/o within               & 80.72            & 80.93            & 65.96              & 64.99             & 75.44          & 74.59          & 64.06          & 63.40          & 40.28                    & 39.63                    & 54.52                    & 53.50                    & 60.86                    & 59.16                   & 42.28                    & 41.79                   \\
w/o across               & 81.18            & 80.19            & 64.40              & 63.57             & 74.67          & 73.68          & 63.08          & 62.56          & 41.88                    & 41.38                    & 54.76                    & 53.92                    & 61.52                    & 61.26                   & 43.84                    & 42.29                   \\
w/o degree               & 79.14            & 80.16            & 63.49              & 62.12             & 74.36          & 73.28          & 63.68          & 62.96          & \color{gray}{30.89}                   & \color{gray}{21.05}                    & \color{gray}{32.82}                    & \color{gray}{23.07}                    & \color{gray}{31.12}                    & \color{gray}{21.61}                   & \color{gray}{30.44}                    & \color{gray}{21.08}                   \\
ours                     & \textbf{82.80}   & \textbf{82.49}   & \textbf{67.30}     & \textbf{66.30}    & \textbf{75.88} & \textbf{75.05} & \textbf{66.34} & \textbf{65.70} & \textbf{42.64}           & \textbf{41.27}           & \textbf{56.36}           & \textbf{55.25}           & \textbf{62.44}           & \textbf{61.66}          & \textbf{46.24}           & \textbf{44.54}          \\ \bottomrule
\end{tabular}%
}
\end{table*}

\begin{figure}
    \centering
    \includegraphics[width=0.23\textwidth]{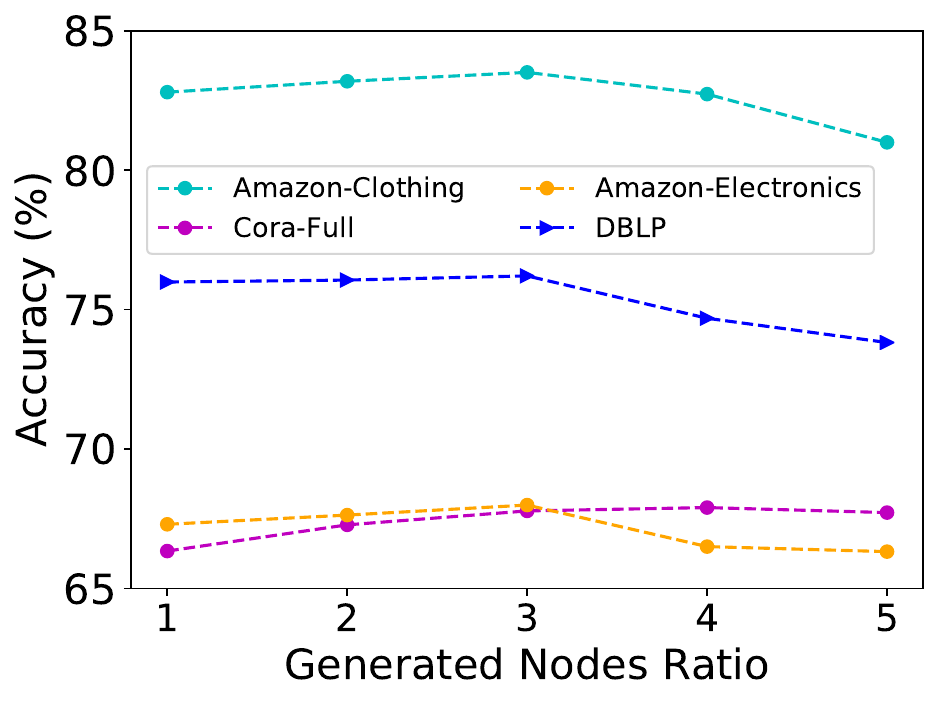}
    \includegraphics[width=0.23\textwidth]{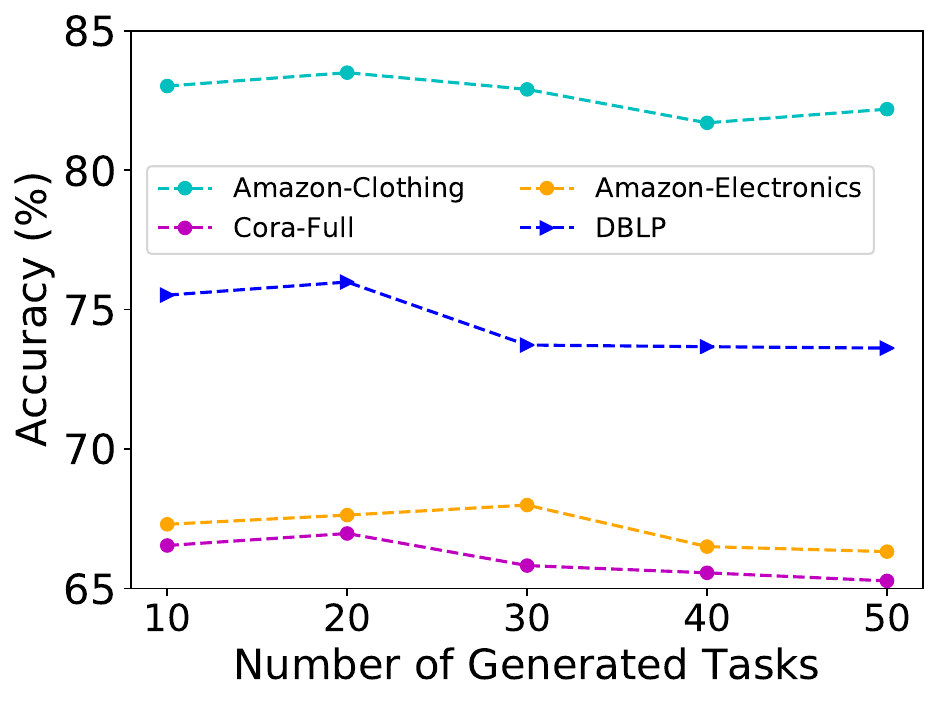}
    \caption{Results vary %across different datasets 
    with hyperparameters.}%Hyperparameter sensitivity on different datasets.
    \label{hyper}
\end{figure}

%%\paragraph{Hyperparameter Sensitivity.} 
\noindent \textbf{Hyperparameter Sensitivity.}
%\subsection{Hyperparameter Sensitivity}
In the 5-way 5-shot in-domain setting, we investigate the impact of two primary hyperparameters on the model performance: the ratio of generated nodes to the original nodes per task (\textit{i.e.}, $\frac{n_{s}^\prime+n_{q}^\prime}{n_s+n_q}$), and the number of generated tasks $\mathrm{T}_{aug}$. Notably, when the studied hyperparameter changes, we set others to their default values. The results are presented in Fig. \ref{hyper}. We can observe that both parameters demonstrate similar trends, with the model performance showing an initial increase followed by a decrease. We attribute this behavior to the substantial enrichment of data diversity by increasing the number of nodes within each task or the number of tasks. However, beyond a certain threshold, the introduced additional data fails to further densify the data distribution, resulting in limited information gain. %Moreover, we also provide the visualization study in \textbf{Appendix} \ref{visualization_study}. %and the time complexity analysis in Appendix \ref{complexity}.
 
\section{Conclusion}
In this work, we propose a simple yet effective approach, called SMILE, for graph few-shot learning with fewer tasks. Specifically, we introduce a novel dual-level mixup strategy, including within-task and across-task mixup, for enriching the diversity of nodes within each task and the diversity of tasks. Also, we incorporate the degree-based prior information to learn expressive node embeddings. Theoretically, we prove that SMILE effectively enhances the model's generalization performance. Empirically, we conduct extensive experiments on multiple benchmarks and the results suggest that SMILE significantly outperforms other baselines, including both in-domain and cross-domain few-shot settings.

\begin{acks}
Our work is supported by the National Science and Technology Major Project under Grant No. 2021ZD0112500, the National Natural Science Foundation of China (No. 62172187 and No. 62372209). Fausto Giunchiglia’s work is funded by European Union’s Horizon 2020 FET Proactive Project (No. 823783).
\end{acks}

\bibliographystyle{ACM-Reference-Format}
\balance
\bibliography{sample-base}


\begin{thebibliography}{59}

%%% the Web address.

\ifx \showCODEN    \undefined \def \showCODEN     #1{\unskip}     \fi
\ifx \showDOI      \undefined \def \showDOI       #1{#1}\fi
\ifx \showISBNx    \undefined \def \showISBNx     #1{\unskip}     \fi
\ifx \showISBNxiii \undefined \def \showISBNxiii  #1{\unskip}     \fi
\ifx \showISSN     \undefined \def \showISSN      #1{\unskip}     \fi
\ifx \showLCCN     \undefined \def \showLCCN      #1{\unskip}     \fi
\ifx \shownote     \undefined \def \shownote      #1{#1}          \fi
\ifx \showarticletitle \undefined \def \showarticletitle #1{#1}   \fi
\ifx \showURL      \undefined \def \showURL       {\relax}        \fi
% invisible to TeX
\providecommand\bibfield[2]{#2}
\providecommand\bibinfo[2]{#2}
\providecommand\natexlab[1]{#1}
\providecommand\showeprint[2][]{arXiv:#2}

\bibitem[Bao et~al\mbox{.}(2020)]%
        {bao2019few}
\bibfield{author}{\bibinfo{person}{Yujia Bao}, \bibinfo{person}{Menghua Wu}, \bibinfo{person}{Shiyu Chang}, {and} \bibinfo{person}{Regina Barzilay}.} \bibinfo{year}{2020}\natexlab{}.
\newblock \showarticletitle{Few-shot text classification with distributional signatures}. In \bibinfo{booktitle}{\emph{ICLR}}.
\newblock


\bibitem[Bateni et~al\mbox{.}(2020)]%
        {bateni2020improved}
\bibfield{author}{\bibinfo{person}{Peyman Bateni}, \bibinfo{person}{Raghav Goyal}, \bibinfo{person}{Vaden Masrani}, \bibinfo{person}{Frank Wood}, {and} \bibinfo{person}{Leonid Sigal}.} \bibinfo{year}{2020}\natexlab{}.
\newblock \showarticletitle{Improved few-shot visual classification}. In \bibinfo{booktitle}{\emph{CVPR}}.
\newblock


\bibitem[Bojchevski and G{\"u}nnemann(2018)]%
        {bojchevski2017deep}
\bibfield{author}{\bibinfo{person}{Aleksandar Bojchevski} {and} \bibinfo{person}{Stephan G{\"u}nnemann}.} \bibinfo{year}{2018}\natexlab{}.
\newblock \showarticletitle{Deep gaussian embedding of graphs: Unsupervised inductive learning via ranking}. In \bibinfo{booktitle}{\emph{ICLR}}.
\newblock


\bibitem[Ding et~al\mbox{.}(2020)]%
        {ding2020graph}
\bibfield{author}{\bibinfo{person}{Kaize Ding}, \bibinfo{person}{Jianling Wang}, \bibinfo{person}{Jundong Li}, \bibinfo{person}{Kai Shu}, \bibinfo{person}{Chenghao Liu}, {and} \bibinfo{person}{Huan Liu}.} \bibinfo{year}{2020}\natexlab{}.
\newblock \showarticletitle{Graph prototypical networks for few-shot learning on attributed networks}. In \bibinfo{booktitle}{\emph{CIKM}}. \bibinfo{pages}{295--304}.
\newblock


\bibitem[Finn et~al\mbox{.}(2017)]%
        {finn2017model}
\bibfield{author}{\bibinfo{person}{Chelsea Finn}, \bibinfo{person}{Pieter Abbeel}, {and} \bibinfo{person}{Sergey Levine}.} \bibinfo{year}{2017}\natexlab{}.
\newblock \showarticletitle{Model-agnostic meta-learning for fast adaptation of deep networks}. In \bibinfo{booktitle}{\emph{ICML}}. \bibinfo{pages}{1126--1135}.
\newblock


\bibitem[Finn et~al\mbox{.}(2019)]%
        {finn2019online}
\bibfield{author}{\bibinfo{person}{Chelsea Finn}, \bibinfo{person}{Aravind Rajeswaran}, \bibinfo{person}{Sham Kakade}, {and} \bibinfo{person}{Sergey Levine}.} \bibinfo{year}{2019}\natexlab{}.
\newblock \showarticletitle{Online meta-learning}. In \bibinfo{booktitle}{\emph{ICML}}.
\newblock


\bibitem[Finn et~al\mbox{.}(2018)]%
        {finn2018probabilistic}
\bibfield{author}{\bibinfo{person}{Chelsea Finn}, \bibinfo{person}{Kelvin Xu}, {and} \bibinfo{person}{Sergey Levine}.} \bibinfo{year}{2018}\natexlab{}.
\newblock \showarticletitle{Probabilistic model-agnostic meta-learning}. In \bibinfo{booktitle}{\emph{NeurIPS}}.
\newblock


\bibitem[Flennerhag et~al\mbox{.}(2020)]%
        {flennerhag2019meta}
\bibfield{author}{\bibinfo{person}{Sebastian Flennerhag}, \bibinfo{person}{Andrei~A Rusu}, \bibinfo{person}{Razvan Pascanu}, \bibinfo{person}{Francesco Visin}, \bibinfo{person}{Hujun Yin}, {and} \bibinfo{person}{Raia Hadsell}.} \bibinfo{year}{2020}\natexlab{}.
\newblock \showarticletitle{Meta-Learning with Warped Gradient Descent}. In \bibinfo{booktitle}{\emph{ICLR}}.
\newblock


\bibitem[Grant et~al\mbox{.}(2019)]%
        {grant2018recasting}
\bibfield{author}{\bibinfo{person}{Erin Grant}, \bibinfo{person}{Chelsea Finn}, \bibinfo{person}{Sergey Levine}, \bibinfo{person}{Trevor Darrell}, {and} \bibinfo{person}{Thomas Griffiths}.} \bibinfo{year}{2019}\natexlab{}.
\newblock \showarticletitle{Recasting gradient-based meta-learning as hierarchical bayes}. In \bibinfo{booktitle}{\emph{ICLR}}.
\newblock


\bibitem[Guan et~al\mbox{.}(2021)]%
        {liu2021vplag}
\bibfield{author}{\bibinfo{person}{Renchu Guan}, \bibinfo{person}{Yonghao Liu}, \bibinfo{person}{Xiaoyue Feng}, {and} \bibinfo{person}{Ximing Li}.} \bibinfo{year}{2021}\natexlab{}.
\newblock \showarticletitle{Paper-publication Prediction with Graph Neural Networks}. In \bibinfo{booktitle}{\emph{CIKM}}.
\newblock


\bibitem[Guo et~al\mbox{.}(2021)]%
        {Guo21few}
\bibfield{author}{\bibinfo{person}{Zhichun Guo}, \bibinfo{person}{Chuxu Zhang}, \bibinfo{person}{Wenhao Yu}, \bibinfo{person}{John Herr}, \bibinfo{person}{Olaf Wiest}, \bibinfo{person}{Meng Jiang}, {and} \bibinfo{person}{Nitesh~V. Chawla}.} \bibinfo{year}{2021}\natexlab{}.
\newblock \showarticletitle{Few-Shot Graph Learning for Molecular Property Prediction}. In \bibinfo{booktitle}{\emph{The Web Conference}}. \bibinfo{pages}{2559--2567}.
\newblock


\bibitem[Hariharan and Girshick(2017)]%
        {hariharan2017low}
\bibfield{author}{\bibinfo{person}{Bharath Hariharan} {and} \bibinfo{person}{Ross Girshick}.} \bibinfo{year}{2017}\natexlab{}.
\newblock \showarticletitle{Low-shot visual recognition by shrinking and hallucinating features}. In \bibinfo{booktitle}{\emph{ICCV}}. \bibinfo{pages}{3018--3027}.
\newblock


\bibitem[Hospedales et~al\mbox{.}(2021)]%
        {hospedales2021meta}
\bibfield{author}{\bibinfo{person}{Timothy Hospedales}, \bibinfo{person}{Antreas Antoniou}, \bibinfo{person}{Paul Micaelli}, {and} \bibinfo{person}{Amos Storkey}.} \bibinfo{year}{2021}\natexlab{}.
\newblock \showarticletitle{Meta-learning in neural networks: A survey}.
\newblock \bibinfo{journal}{\emph{IEEE TPAMI}} \bibinfo{volume}{44}, \bibinfo{number}{9} (\bibinfo{year}{2021}), \bibinfo{pages}{5149--5169}.
\newblock


\bibitem[Huang and Zitnik(2020)]%
        {huang2020graph}
\bibfield{author}{\bibinfo{person}{Kexin Huang} {and} \bibinfo{person}{Marinka Zitnik}.} \bibinfo{year}{2020}\natexlab{}.
\newblock \showarticletitle{Graph meta learning via local subgraphs}. In \bibinfo{booktitle}{\emph{NeurIPS}}. \bibinfo{pages}{5862--5874}.
\newblock


\bibitem[Jamal and Qi(2019)]%
        {jamal2019task}
\bibfield{author}{\bibinfo{person}{Muhammad~Abdullah Jamal} {and} \bibinfo{person}{Guo-Jun Qi}.} \bibinfo{year}{2019}\natexlab{}.
\newblock \showarticletitle{Task agnostic meta-learning for few-shot learning}. In \bibinfo{booktitle}{\emph{CVPR}}. \bibinfo{pages}{11719--11727}.
\newblock


\bibitem[Kim et~al\mbox{.}(2023)]%
        {kim2023task}
\bibfield{author}{\bibinfo{person}{Sungwon Kim}, \bibinfo{person}{Junseok Lee}, \bibinfo{person}{Namkyeong Lee}, \bibinfo{person}{Wonjoong Kim}, \bibinfo{person}{Seungyoon Choi}, {and} \bibinfo{person}{Chanyoung Park}.} \bibinfo{year}{2023}\natexlab{}.
\newblock \showarticletitle{Task-Equivariant Graph Few-shot Learning}. In \bibinfo{booktitle}{\emph{SIGKDD}}.
\newblock


\bibitem[Kipf and Welling(2017)]%
        {kipf2016semi}
\bibfield{author}{\bibinfo{person}{Thomas~N Kipf} {and} \bibinfo{person}{Max Welling}.} \bibinfo{year}{2017}\natexlab{}.
\newblock \showarticletitle{Semi-supervised classification with graph convolutional networks}. In \bibinfo{booktitle}{\emph{ICLR}}.
\newblock


\bibitem[Lee et~al\mbox{.}(2022)]%
        {lee2022set}
\bibfield{author}{\bibinfo{person}{Seanie Lee}, \bibinfo{person}{Bruno Andreis}, \bibinfo{person}{Kenji Kawaguchi}, \bibinfo{person}{Juho Lee}, {and} \bibinfo{person}{Sung~Ju Hwang}.} \bibinfo{year}{2022}\natexlab{}.
\newblock \showarticletitle{Set-based meta-interpolation for few-task meta-learning}. In \bibinfo{booktitle}{\emph{NeurIPS}}. \bibinfo{pages}{6775--6788}.
\newblock


\bibitem[Li et~al\mbox{.}(2024)]%
        {li2024simple}
\bibfield{author}{\bibinfo{person}{Mengyu Li}, \bibinfo{person}{Yonghao Liu}, \bibinfo{person}{Fausto Giunchiglia}, \bibinfo{person}{Xiaoyue Feng}, {and} \bibinfo{person}{Renchu Guan}.} \bibinfo{year}{2024}\natexlab{}.
\newblock \showarticletitle{Simple-Sampling and Hard-Mixup with Prototypes to Rebalance Contrastive Learning for Text Classification}.
\newblock \bibinfo{journal}{\emph{arxiv preprint arXiv:2405.11524}} (\bibinfo{year}{2024}).
\newblock


\bibitem[Lifchitz et~al\mbox{.}(2019)]%
        {lifchitz2019dense}
\bibfield{author}{\bibinfo{person}{Yann Lifchitz}, \bibinfo{person}{Yannis Avrithis}, \bibinfo{person}{Sylvaine Picard}, {and} \bibinfo{person}{Andrei Bursuc}.} \bibinfo{year}{2019}\natexlab{}.
\newblock \showarticletitle{Dense classification and implanting for few-shot learning}. In \bibinfo{booktitle}{\emph{CVPR}}.
\newblock


\bibitem[Liu et~al\mbox{.}(2019)]%
        {liu2019learning}
\bibfield{author}{\bibinfo{person}{Lu Liu}, \bibinfo{person}{Tianyi Zhou}, \bibinfo{person}{Guodong Long}, \bibinfo{person}{Jing Jiang}, {and} \bibinfo{person}{Chengqi Zhang}.} \bibinfo{year}{2019}\natexlab{}.
\newblock \showarticletitle{Learning to propagate for graph meta-learning}. In \bibinfo{booktitle}{\emph{NeurIPS}}.
\newblock


\bibitem[Liu et~al\mbox{.}(2025a)]%
        {liu2025improved}
\bibfield{author}{\bibinfo{person}{Yonghao Liu}, \bibinfo{person}{Fausto Giunchiglia}, \bibinfo{person}{Lan Huang}, \bibinfo{person}{Ximing Li}, \bibinfo{person}{Xiaoyue Feng}, {and} \bibinfo{person}{Renchu Guan}.} \bibinfo{year}{2025}\natexlab{a}.
\newblock \showarticletitle{A Simple Graph Contrastive Learning Framework for Short Text Classification}. In \bibinfo{booktitle}{\emph{AAAI}}.
\newblock


\bibitem[Liu et~al\mbox{.}(2025b)]%
        {liu2025enhancing}
\bibfield{author}{\bibinfo{person}{Yonghao Liu}, \bibinfo{person}{Fausto Giunchiglia}, \bibinfo{person}{Ximing Li}, \bibinfo{person}{Lan Huang}, \bibinfo{person}{Xiaoyue Feng}, {and} \bibinfo{person}{Renchu Guan}.} \bibinfo{year}{2025}\natexlab{b}.
\newblock \showarticletitle{Enhancing Unsupervised Graph Few-shot Learning via Set Functions and Optimal Transport}. In \bibinfo{booktitle}{\emph{SIGKDD}}.
\newblock


\bibitem[Liu et~al\mbox{.}(2021)]%
        {liu2021deep}
\bibfield{author}{\bibinfo{person}{Yonghao Liu}, \bibinfo{person}{Renchu Guan}, \bibinfo{person}{Fausto Giunchiglia}, \bibinfo{person}{Yanchun Liang}, {and} \bibinfo{person}{Xiaoyue Feng}.} \bibinfo{year}{2021}\natexlab{}.
\newblock \showarticletitle{Deep attention diffusion graph neural networks for text classification}. In \bibinfo{booktitle}{\emph{EMNLP}}.
\newblock


\bibitem[Liu et~al\mbox{.}(2024a)]%
        {liu2024simple}
\bibfield{author}{\bibinfo{person}{Yonghao Liu}, \bibinfo{person}{Lan Huang}, \bibinfo{person}{Bowen Cao}, \bibinfo{person}{Ximing Li}, \bibinfo{person}{Fausto Giunchiglia}, \bibinfo{person}{Xiaoyue Feng}, {and} \bibinfo{person}{Renchu Guan}.} \bibinfo{year}{2024}\natexlab{a}.
\newblock \showarticletitle{A Simple but Effective Approach for Unsupervised Few-Shot Graph Classification}. In \bibinfo{booktitle}{\emph{WWW}}.
\newblock


\bibitem[Liu et~al\mbox{.}(2024b)]%
        {liu2024improved}
\bibfield{author}{\bibinfo{person}{Yonghao Liu}, \bibinfo{person}{Lan Huang}, \bibinfo{person}{Fausto Giunchiglia}, \bibinfo{person}{Xiaoyue Feng}, {and} \bibinfo{person}{Renchu Guan}.} \bibinfo{year}{2024}\natexlab{b}.
\newblock \showarticletitle{Improved Graph Contrastive Learning for Short Text Classification}. In \bibinfo{booktitle}{\emph{AAAI}}.
\newblock


\bibitem[Liu et~al\mbox{.}(2022)]%
        {liu2022few}
\bibfield{author}{\bibinfo{person}{Yonghao Liu}, \bibinfo{person}{Mengyu Li}, \bibinfo{person}{Ximing Li}, \bibinfo{person}{Fausto Giunchiglia}, \bibinfo{person}{Xiaoyue Feng}, {and} \bibinfo{person}{Renchu Guan}.} \bibinfo{year}{2022}\natexlab{}.
\newblock \showarticletitle{Few-shot node classification on attributed networks with graph meta-learning}. In \bibinfo{booktitle}{\emph{SIGIR}}.
\newblock


\bibitem[Liu et~al\mbox{.}(2024c)]%
        {liu2024meta}
\bibfield{author}{\bibinfo{person}{Yonghao Liu}, \bibinfo{person}{Mengyu Li}, \bibinfo{person}{Ximing Li}, \bibinfo{person}{Lan Huang}, \bibinfo{person}{Fausto Giunchiglia}, \bibinfo{person}{Yanchun Liang}, \bibinfo{person}{Xiaoyue Feng}, {and} \bibinfo{person}{Renchu Guan}.} \bibinfo{year}{2024}\natexlab{c}.
\newblock \showarticletitle{Meta-GPS++: Enhancing Graph Meta-Learning with Contrastive Learning and Self-Training}.
\newblock \bibinfo{journal}{\emph{ACM TKDD}} \bibinfo{volume}{18}, \bibinfo{number}{9} (\bibinfo{year}{2024}), \bibinfo{pages}{1--30}.
\newblock


\bibitem[Liu et~al\mbox{.}(2024d)]%
        {liu2024resolving}
\bibfield{author}{\bibinfo{person}{Yonghao Liu}, \bibinfo{person}{Mengyu Li}, \bibinfo{person}{Di Liang}, \bibinfo{person}{Ximing Li}, \bibinfo{person}{Fausto Giunchiglia}, \bibinfo{person}{Lan Huang}, \bibinfo{person}{Xiaoyue Feng}, {and} \bibinfo{person}{Renchu Guan}.} \bibinfo{year}{2024}\natexlab{d}.
\newblock \showarticletitle{Resolving Word Vagueness with Scenario-guided Adapter for Natural Language Inference}. In \bibinfo{booktitle}{\emph{IJCAI}}.
\newblock


\bibitem[Liu et~al\mbox{.}(2025c)]%
        {liu2025boosting}
\bibfield{author}{\bibinfo{person}{Yonghao Liu}, \bibinfo{person}{Mengyu Li}, \bibinfo{person}{Wei Pang}, \bibinfo{person}{Fausto Giunchiglia}, \bibinfo{person}{Lan Huang}, \bibinfo{person}{Xiaoyue Feng}, {and} \bibinfo{person}{Renchu Guan}.} \bibinfo{year}{2025}\natexlab{c}.
\newblock \showarticletitle{Boosting Short Text Classification with Multi-Source Information Exploration and Dual-Level Contrastive Learning}. In \bibinfo{booktitle}{\emph{AAAI}}.
\newblock


\bibitem[Liu et~al\mbox{.}(2023a)]%
        {liu2023time}
\bibfield{author}{\bibinfo{person}{Yonghao Liu}, \bibinfo{person}{Di Liang}, \bibinfo{person}{Fang Fang}, \bibinfo{person}{Sirui Wang}, \bibinfo{person}{Wei Wu}, {and} \bibinfo{person}{Rui Jiang}.} \bibinfo{year}{2023}\natexlab{a}.
\newblock \showarticletitle{Time-aware multiway adaptive fusion network for temporal knowledge graph question answering}. In \bibinfo{booktitle}{\emph{ICASSP}}. \bibinfo{pages}{1--5}.
\newblock


\bibitem[Liu et~al\mbox{.}(2023b)]%
        {liulocal}
\bibfield{author}{\bibinfo{person}{Yonghao Liu}, \bibinfo{person}{Di Liang}, \bibinfo{person}{Mengyu Li}, \bibinfo{person}{Fausto Giunchiglia}, \bibinfo{person}{Ximing Li}, \bibinfo{person}{Sirui Wang}, \bibinfo{person}{Wei Wu}, \bibinfo{person}{Lan Huang}, \bibinfo{person}{Xiaoyue Feng}, {and} \bibinfo{person}{Renchu Guan}.} \bibinfo{year}{2023}\natexlab{b}.
\newblock \showarticletitle{Local and Global: Temporal Question Answering via Information Fusion}. In \bibinfo{booktitle}{\emph{IJCAI}}.
\newblock


\bibitem[Livingstone(2000)]%
        {livingstone2000characterization}
\bibfield{author}{\bibinfo{person}{David~J Livingstone}.} \bibinfo{year}{2000}\natexlab{}.
\newblock \showarticletitle{The characterization of chemical structures using molecular properties. A survey}.
\newblock \bibinfo{journal}{\emph{Journal of Chemical Information and Computer Sciences}} \bibinfo{volume}{40}, \bibinfo{number}{2} (\bibinfo{year}{2000}), \bibinfo{pages}{195--209}.
\newblock


\bibitem[McAuley et~al\mbox{.}(2015)]%
        {mcauley2015inferring}
\bibfield{author}{\bibinfo{person}{Julian McAuley}, \bibinfo{person}{Rahul Pandey}, {and} \bibinfo{person}{Jure Leskovec}.} \bibinfo{year}{2015}\natexlab{}.
\newblock \showarticletitle{Inferring networks of substitutable and complementary products}. In \bibinfo{booktitle}{\emph{SIGKDD}}. \bibinfo{pages}{785--794}.
\newblock


\bibitem[Mishra et~al\mbox{.}(2018)]%
        {mishra2017simple}
\bibfield{author}{\bibinfo{person}{Nikhil Mishra}, \bibinfo{person}{Mostafa Rohaninejad}, \bibinfo{person}{Xi Chen}, {and} \bibinfo{person}{Pieter Abbeel}.} \bibinfo{year}{2018}\natexlab{}.
\newblock \showarticletitle{A simple neural attentive meta-learner}. In \bibinfo{booktitle}{\emph{ICLR}}.
\newblock


\bibitem[Mukherjee and Awadallah(2020)]%
        {mukherjee2020uncertainty}
\bibfield{author}{\bibinfo{person}{Subhabrata Mukherjee} {and} \bibinfo{person}{Ahmed Awadallah}.} \bibinfo{year}{2020}\natexlab{}.
\newblock \showarticletitle{Uncertainty-aware self-training for few-shot text classification}. In \bibinfo{booktitle}{\emph{NeurIPS}}. \bibinfo{pages}{21199--21212}.
\newblock


\bibitem[Ni et~al\mbox{.}(2021)]%
        {ni2021data}
\bibfield{author}{\bibinfo{person}{Renkun Ni}, \bibinfo{person}{Micah Goldblum}, \bibinfo{person}{Amr Sharaf}, \bibinfo{person}{Kezhi Kong}, {and} \bibinfo{person}{Tom Goldstein}.} \bibinfo{year}{2021}\natexlab{}.
\newblock \showarticletitle{Data augmentation for meta-learning}. In \bibinfo{booktitle}{\emph{ICML}}. \bibinfo{pages}{8152--8161}.
\newblock


\bibitem[Oh et~al\mbox{.}(2021)]%
        {oh2020boil}
\bibfield{author}{\bibinfo{person}{Jaehoon Oh}, \bibinfo{person}{Hyungjun Yoo}, \bibinfo{person}{ChangHwan Kim}, {and} \bibinfo{person}{Se-Young Yun}.} \bibinfo{year}{2021}\natexlab{}.
\newblock \showarticletitle{BOIL: Towards Representation Change for Few-shot Learning}. In \bibinfo{booktitle}{\emph{ICLR}}.
\newblock


\bibitem[Park et~al\mbox{.}(2019)]%
        {Park2019estimating}
\bibfield{author}{\bibinfo{person}{Namyong Park}, \bibinfo{person}{Andrey Kan}, \bibinfo{person}{Xin~Luna Dong}, \bibinfo{person}{Tong Zhao}, {and} \bibinfo{person}{Christos Faloutsos}.} \bibinfo{year}{2019}\natexlab{}.
\newblock \showarticletitle{Estimating Node Importance in Knowledge Graphs Using Graph Neural Networks}. In \bibinfo{booktitle}{\emph{SIGKDD}}. \bibinfo{pages}{596--606}.
\newblock


\bibitem[Rajendran et~al\mbox{.}(2020)]%
        {rajendran2020meta}
\bibfield{author}{\bibinfo{person}{Janarthanan Rajendran}, \bibinfo{person}{Alexander Irpan}, {and} \bibinfo{person}{Eric Jang}.} \bibinfo{year}{2020}\natexlab{}.
\newblock \showarticletitle{Meta-learning requires meta-augmentation}. In \bibinfo{booktitle}{\emph{NeurIPS}}. \bibinfo{pages}{5705--5715}.
\newblock


\bibitem[Rusu et~al\mbox{.}(2019)]%
        {rusu2018meta}
\bibfield{author}{\bibinfo{person}{Andrei~A Rusu}, \bibinfo{person}{Dushyant Rao}, \bibinfo{person}{Jakub Sygnowski}, \bibinfo{person}{Oriol Vinyals}, \bibinfo{person}{Razvan Pascanu}, \bibinfo{person}{Simon Osindero}, {and} \bibinfo{person}{Raia Hadsell}.} \bibinfo{year}{2019}\natexlab{}.
\newblock \showarticletitle{Meta-learning with latent embedding optimization}. In \bibinfo{booktitle}{\emph{ICLR}}.
\newblock


\bibitem[Snell et~al\mbox{.}(2017)]%
        {snell2017prototypical}
\bibfield{author}{\bibinfo{person}{Jake Snell}, \bibinfo{person}{Kevin Swersky}, {and} \bibinfo{person}{Richard Zemel}.} \bibinfo{year}{2017}\natexlab{}.
\newblock \showarticletitle{Prototypical networks for few-shot learning}. In \bibinfo{booktitle}{\emph{NeurIPS}}.
\newblock


\bibitem[Tan et~al\mbox{.}(2023)]%
        {tan2023virtual}
\bibfield{author}{\bibinfo{person}{Zhen Tan}, \bibinfo{person}{Ruocheng Guo}, \bibinfo{person}{Kaize Ding}, {and} \bibinfo{person}{Huan Liu}.} \bibinfo{year}{2023}\natexlab{}.
\newblock \showarticletitle{Virtual node tuning for few-shot node classification}. In \bibinfo{booktitle}{\emph{SIGKDD}}.
\newblock


\bibitem[Tan et~al\mbox{.}(2022)]%
        {tan2022transductive}
\bibfield{author}{\bibinfo{person}{Zhen Tan}, \bibinfo{person}{Song Wang}, \bibinfo{person}{Kaize Ding}, \bibinfo{person}{Jundong Li}, {and} \bibinfo{person}{Huan Liu}.} \bibinfo{year}{2022}\natexlab{}.
\newblock \showarticletitle{Transductive Linear Probing: A Novel Framework for Few-Shot Node Classification}. In \bibinfo{booktitle}{\emph{LoG}}. \bibinfo{pages}{4--1}.
\newblock


\bibitem[Tang et~al\mbox{.}(2008)]%
        {tang2008arnetminer}
\bibfield{author}{\bibinfo{person}{Jie Tang}, \bibinfo{person}{Jing Zhang}, \bibinfo{person}{Limin Yao}, \bibinfo{person}{Juanzi Li}, \bibinfo{person}{Li Zhang}, {and} \bibinfo{person}{Zhong Su}.} \bibinfo{year}{2008}\natexlab{}.
\newblock \showarticletitle{Arnetminer: extraction and mining of academic social networks}. In \bibinfo{booktitle}{\emph{SIGKDD}}. \bibinfo{pages}{990--998}.
\newblock


\bibitem[Tian et~al\mbox{.}(2020)]%
        {tian2020rethinking}
\bibfield{author}{\bibinfo{person}{Yonglong Tian}, \bibinfo{person}{Yue Wang}, \bibinfo{person}{Dilip Krishnan}, \bibinfo{person}{Joshua~B Tenenbaum}, {and} \bibinfo{person}{Phillip Isola}.} \bibinfo{year}{2020}\natexlab{}.
\newblock \showarticletitle{Rethinking few-shot image classification: a good embedding is all you need?}. In \bibinfo{booktitle}{\emph{ECCV}}. \bibinfo{pages}{266--282}.
\newblock


\bibitem[Vinyals et~al\mbox{.}(2016)]%
        {vinyals2016matching}
\bibfield{author}{\bibinfo{person}{Oriol Vinyals}, \bibinfo{person}{Charles Blundell}, \bibinfo{person}{Timothy Lillicrap}, \bibinfo{person}{Daan Wierstra}, {et~al\mbox{.}}} \bibinfo{year}{2016}\natexlab{}.
\newblock \showarticletitle{Matching networks for one shot learning}. In \bibinfo{booktitle}{\emph{NeurIPS}}.
\newblock


\bibitem[Wang et~al\mbox{.}(2021)]%
        {wang2021grad2task}
\bibfield{author}{\bibinfo{person}{Jixuan Wang}, \bibinfo{person}{Kuan-Chieh Wang}, \bibinfo{person}{Frank Rudzicz}, {and} \bibinfo{person}{Michael Brudno}.} \bibinfo{year}{2021}\natexlab{}.
\newblock \showarticletitle{Grad2task: Improved few-shot text classification using gradients for task representation}. In \bibinfo{booktitle}{\emph{NeurIPS}}.
\newblock


\bibitem[Wang et~al\mbox{.}(2022)]%
        {wang2022task}
\bibfield{author}{\bibinfo{person}{Song Wang}, \bibinfo{person}{Kaize Ding}, \bibinfo{person}{Chuxu Zhang}, \bibinfo{person}{Chen Chen}, {and} \bibinfo{person}{Jundong Li}.} \bibinfo{year}{2022}\natexlab{}.
\newblock \showarticletitle{Task-adaptive few-shot node classification}. In \bibinfo{booktitle}{\emph{SIGKDD}}. \bibinfo{pages}{1910--1919}.
\newblock


\bibitem[Wang et~al\mbox{.}(2023a)]%
        {wang2023few}
\bibfield{author}{\bibinfo{person}{Song Wang}, \bibinfo{person}{Yushun Dong}, \bibinfo{person}{Kaize Ding}, \bibinfo{person}{Chen Chen}, {and} \bibinfo{person}{Jundong Li}.} \bibinfo{year}{2023}\natexlab{a}.
\newblock \showarticletitle{Few-shot node classification with extremely weak supervision}. In \bibinfo{booktitle}{\emph{WSDM}}. \bibinfo{pages}{276--284}.
\newblock


\bibitem[Wang et~al\mbox{.}(2023b)]%
        {wang2023contrastive}
\bibfield{author}{\bibinfo{person}{Song Wang}, \bibinfo{person}{Zhen Tan}, \bibinfo{person}{Huan Liu}, {and} \bibinfo{person}{Jundong Li}.} \bibinfo{year}{2023}\natexlab{b}.
\newblock \showarticletitle{Contrastive Meta-Learning for Few-shot Node Classification}. In \bibinfo{booktitle}{\emph{SIGKDD}}.
\newblock


\bibitem[Wu et~al\mbox{.}(2019)]%
        {wu2019simplifying}
\bibfield{author}{\bibinfo{person}{Felix Wu}, \bibinfo{person}{Amauri Souza}, \bibinfo{person}{Tianyi Zhang}, \bibinfo{person}{Christopher Fifty}, \bibinfo{person}{Tao Yu}, {and} \bibinfo{person}{Kilian Weinberger}.} \bibinfo{year}{2019}\natexlab{}.
\newblock \showarticletitle{Simplifying graph convolutional networks}. In \bibinfo{booktitle}{\emph{ICML}}.
\newblock


\bibitem[Xu et~al\mbox{.}(2021)]%
        {xu2021augnlg}
\bibfield{author}{\bibinfo{person}{Xinnuo Xu}, \bibinfo{person}{Guoyin Wang}, \bibinfo{person}{Young-Bum Kim}, {and} \bibinfo{person}{Sungjin Lee}.} \bibinfo{year}{2021}\natexlab{}.
\newblock \showarticletitle{AugNLG: Few-shot Natural Language Generation using Self-trained Data Augmentation}. In \bibinfo{booktitle}{\emph{ACL}}.
\newblock


\bibitem[Yao et~al\mbox{.}(2021)]%
        {yao2021improving}
\bibfield{author}{\bibinfo{person}{Huaxiu Yao}, \bibinfo{person}{Long-Kai Huang}, \bibinfo{person}{Linjun Zhang}, \bibinfo{person}{Ying Wei}, \bibinfo{person}{Li Tian}, \bibinfo{person}{James Zou}, \bibinfo{person}{Junzhou Huang}, {et~al\mbox{.}}} \bibinfo{year}{2021}\natexlab{}.
\newblock \showarticletitle{Improving generalization in meta-learning via task augmentation}. In \bibinfo{booktitle}{\emph{ICML}}. \bibinfo{pages}{11887--11897}.
\newblock


\bibitem[Yao et~al\mbox{.}(2022)]%
        {yao2021meta}
\bibfield{author}{\bibinfo{person}{Huaxiu Yao}, \bibinfo{person}{Linjun Zhang}, {and} \bibinfo{person}{Chelsea Finn}.} \bibinfo{year}{2022}\natexlab{}.
\newblock \showarticletitle{Meta-learning with fewer tasks through task interpolation}. In \bibinfo{booktitle}{\emph{ICLR}}.
\newblock


\bibitem[Yin et~al\mbox{.}(2020)]%
        {yin2019meta}
\bibfield{author}{\bibinfo{person}{Mingzhang Yin}, \bibinfo{person}{George Tucker}, \bibinfo{person}{Mingyuan Zhou}, \bibinfo{person}{Sergey Levine}, {and} \bibinfo{person}{Chelsea Finn}.} \bibinfo{year}{2020}\natexlab{}.
\newblock \showarticletitle{Meta-learning without memorization}. In \bibinfo{booktitle}{\emph{ICLR}}.
\newblock


\bibitem[Zhang et~al\mbox{.}(2022)]%
        {zhang2022few}
\bibfield{author}{\bibinfo{person}{Chuxu Zhang}, \bibinfo{person}{Kaize Ding}, \bibinfo{person}{Jundong Li}, \bibinfo{person}{Xiangliang Zhang}, \bibinfo{person}{Yanfang Ye}, \bibinfo{person}{Nitesh~V Chawla}, {and} \bibinfo{person}{Huan Liu}.} \bibinfo{year}{2022}\natexlab{}.
\newblock \showarticletitle{Few-shot learning on graphs}. In \bibinfo{booktitle}{\emph{IJCAI}}.
\newblock


\bibitem[Zhang et~al\mbox{.}(2019)]%
        {zhang2019variational}
\bibfield{author}{\bibinfo{person}{Jian Zhang}, \bibinfo{person}{Chenglong Zhao}, \bibinfo{person}{Bingbing Ni}, \bibinfo{person}{Minghao Xu}, {and} \bibinfo{person}{Xiaokang Yang}.} \bibinfo{year}{2019}\natexlab{}.
\newblock \showarticletitle{Variational Few-Shot Learning}. In \bibinfo{booktitle}{\emph{ICCV}}. \bibinfo{pages}{1685--1694}.
\newblock


\bibitem[Zhou et~al\mbox{.}(2019)]%
        {zhou2019meta}
\bibfield{author}{\bibinfo{person}{Fan Zhou}, \bibinfo{person}{Chengtai Cao}, \bibinfo{person}{Kunpeng Zhang}, \bibinfo{person}{Goce Trajcevski}, \bibinfo{person}{Ting Zhong}, {and} \bibinfo{person}{Ji Geng}.} \bibinfo{year}{2019}\natexlab{}.
\newblock \showarticletitle{Meta-gnn: On few-shot node classification in graph meta-learning}. In \bibinfo{booktitle}{\emph{CIKM}}. \bibinfo{pages}{2357--2360}.
\newblock


\end{thebibliography}
\vfill\eject

%\onecolumn
\newpage
\appendix
\section*{Appendix}
\section{Description of Symbols}
\label{symbol}

We summarize the used important symbols in Table \ref{symbols}.

\begin{table}[!ht]
\renewcommand{\thetable}{S1}
\centering
\caption{Descriptions of the symbols.}
\label{symbols}
\resizebox{0.5\textwidth}{!}{%
\begin{tabular}{@{}c|c@{}}
\toprule
Symbols & Descriptions \\ \midrule
    $\mathcal{G}, \mathcal{V}, \mathcal{E}$    & Graph, node set, and edge set             \\
    $\mathrm{Z}, \mathrm{A}$    & Initialized node features and adjacency matrix             \\
    $\hat{\mathrm{D}}, \mathrm{H}, \mathrm{X}$    &  Degree matrix, hidden vectors, and refined vectors            \\
    $\kappa, \alpha, \beta$    &   Interaction weights, node centralities, and adjusted scores           \\
    $N, K, M$    & $N$ way, $K$ shot, $M$ query \\
    $\mathcal{D}_{org}$    &   Original meta-training tasks           \\ 
    $\mathcal{D}_{aug}$ & Generated meta-training tasks \\ 
    $\mathcal{D}_{all}$ & All original and generated meta-training tasks  \\
    $\mathcal{S}_t, \mathcal{Q}_t$ & Support and query set \\
    $n_s, n_q$ & Number of samples in $\mathcal{S}_t$ and $\mathcal{Q}_t$ \\
    $\mathcal{T}_{tes}$ & Meta-testing task \\
    $\mathcal{S}_{tes}, \mathcal{Q}_{tes}$ & Support and query set of $\mathcal{T}_{tes}$ \\
    $\eta, \zeta$ & Hyperparameters in Beta distribution \\
    $\lambda$ & Random variable drawn from Beta distribution \\
    $\mathcal{S}_t^\prime, \mathcal{Q}_t^\prime$ & Generated support and query set \\
    $n_{s^\prime}, n_{q^\prime}$ & Number of samples in $\mathcal{S}_t^\prime$ and $\mathcal{Q}_t^\prime$ \\
    $m^\prime, m$ & Number of samples in $\mathcal{S}_t\cup\mathcal{S}_t^\prime$ and $\mathcal{Q}_t\cup\mathcal{Q}_t^\prime$ \\
    $\mathcal{T}_t^{aug}, \tilde{\mathcal{S}}, \tilde{\mathcal{Q}}$ & Interpolated task with its support and query set \\
    $\mathrm{T}_{org}$ & Number of tasks in $\mathcal{D}_{org}$ \\
    $\mathrm{T}_{aug}$ & Number of tasks in $\mathcal{D}_{aug}$ \\
    $\mathrm{T}$ & Number of tasks in $\mathcal{D}_{all}$ \\
    \bottomrule
\end{tabular} %
}
\end{table}

\section{Training Procedure}
We present the training procedure of the proposed SMILE in Algorithm \ref{pseudo}.

\label{training_procedure}
\begin{algorithm}[ht]
\renewcommand{\algorithmicrequire}{\textbf{Input:}}
\renewcommand{\algorithmicensure}{\textbf{Output:}}
\caption{The process of SMILE}
\label{pseudo}
\begin{algorithmic}[1]
\REQUIRE A graph $\mathcal{G}\!=\!\{\mathcal{V},\mathcal{E},\mathrm{Z},\mathrm{A}\}$.\\
\ENSURE The well-trained SMILE.
\STATE // \textit{Meta-training process}
    \WHILE{\textit{not convergence}}
    \STATE Learn node embeddings using Eq.\ref{sgc}.
    \STATE Refine node embeddings using Eq.\ref{refine}.
    \STATE Construct meta-training tasks $\mathcal{D}_{org}$. %and meta-testing task $\mathcal{T}_{tes}$.
    \STATE Perform within-task mixup to obtain the augmented task $\mathcal{T}_t$ using Eq.\ref{intra}.
    \STATE Perform across-task mixup to obtain the interpolated task $\mathcal{T}_t^{aug}$ using Eqs.\ref{prototype} and \ref{proto_mix}.
    \STATE Form the interpolated tasks $\mathcal{D}_{aug}$.
    \STATE Obtain the enriched meta-training tasks $\mathcal{D}_{all}$.
    \STATE Compute the prototypes of support set for each task using Eq.\ref{prototype}.
    \STATE Optimize the model using Eq.\ref{proto}.
    \ENDWHILE
    \STATE // \textit{Meta-testing process}
    \STATE Construct meta-testing task $\mathcal{T}_{tes}$.
    \STATE Compute the prototypes in $\mathcal{S}_{tes}$ %and  predict the node labels in $\mathcal{Q}_{tes}$ 
    using Eq.\ref{meta-test}.  %Fine-tune the model using $\mathcal{S}_{tes}$.
    \STATE Predict the node labels in $\mathcal{Q}_{tes}$. %using the fine-tuned model. 
\end{algorithmic}
\end{algorithm}

\section{Complexity Analysis}
\label{complexity}
We analyze the time complexity of our proposed model to demonstrate its effectiveness. Our model mainly contains two parts, including node presentation learning and dual-level mixup. As linear interpolation is employed in the dual-level mixup, it does not introduce additional time complexity. Basically, most of the time-consuming operations arise from the node embedding process. Here, we choose SGC as the base graph encoder, which removes layer-wise non-linear operations and performs feature extraction in a parameter-free manner. The required time complexity of this step is $O(n^2d)$, where $n$ and $d$ denote the number of nodes and the dimension of node features, respectively. Note that as feature extraction does not require any weights, it is essentially equivalent to a preprocessing step and can be precomputed in practice. Moreover, in the procedure of incorporating degree-based prior information to obtain the refined node representations, the required time complexity is $O(2nd+n)$. Thus, the overall time complexity of our approach is $O(n^2d) + O(2nd+n)$, which is acceptable to us.

\section{Statistics and Descriptions of Datasets}
\label{dataset_description}
In this section, we provide detailed statistics and descriptions of the used datasets, which have been widely used in previous studies \cite{ding2020graph, liu2022few, wang2022task}. The detailed descriptions are provided below.

\noindent $\bullet$ \textbf{Amazon-Clothing} \cite{mcauley2015inferring}: It is a product network constructed from the ``Clothing, Shoes, and Jewelry'' category on Amazon. In this dataset, each product is treated as a node, and its description is used to construct node features. A link is created between products if they are co-viewed. The labels are defined as the low-level product class. For this dataset, we use the 40/17/20 class split for meta-training/meta-validation/meta-testing.

\noindent $\bullet$ \textbf{CoraFull} \cite{bojchevski2017deep}: It is a prevalent citation network. Each node represents a paper, and an edge is created between two papers if one cites the other. The nodes are labeled based on the topics of the papers. This dataset extends the previously widely used small dataset Cora by extracting raw data from the entire network. For this dataset, we use a 25/20/25 node class split for meta-training/meta-validation/meta-testing.

\noindent $\bullet$ \textbf{Amazon-Electronics} \cite{mcauley2015inferring}: It is another Amazon product network that contains products belonging to the ``Electronics" category. Each node represents a product, with its features representing the product description. An edge is created between products if there is a co-purchasing relationship. The low-level product categories are used as class labels. For this dataset, we use a 90/37/40 node category split for meta-training/meta-validation/meta-testing.

\noindent $\bullet$ \textbf{DBLP} \cite{tang2008arnetminer}: It is a citation network where each node represents a paper, and the edges represent citation relationships between different papers. The abstracts of the papers are used to construct node features. The class labels of the nodes are defined as the publication venues of the papers. For this dataset, we use an 80/27/30 node category split for meta-training/meta-validation/meta-testing.

\section{Descriptions of Baselines}
\label{baseline}
In this section, we present the detailed descriptions of the selected baselines below.

\subsection{Traditional Meta-learning Method}
\noindent \textbf{Protonet} \cite{snell2017prototypical}: It learns a metric space by acquiring prototypes of different categories from the support set and computes the similarity between the query samples and each prototype to predict their categories.

\noindent \textbf{MAML} \cite{finn2017model}: It enables the meta-trainer to obtain a well-initialized parameter by performing one or more gradient update steps on the model parameters, allowing for rapid adaptation to downstream novel tasks with limited labeled data.

\subsection{Meta-learning with Fewer Tasks Method}
\noindent \textbf{MetaMix} \cite{yao2021improving}: It enhances meta-training tasks by linearly combining the features and labels of samples from the support and query sets to improve the generalization of the model.

\noindent \textbf{MLTI} \cite{yao2021meta}: It generates additional tasks by randomly sampling a pair of tasks and interpolating their corresponding features and labels, replacing the original tasks for training.

\noindent \textbf{Meta-Inter} \cite{lee2022set}: It proposes a domain-agnostic task augmentation method that utilizes expressive neural set functions to densify the distribution of meta-training tasks through a bi-level optimization process.

\subsection{Graph Meta-learning Method}
\noindent \textbf{Meta-GNN} \cite{zhou2019meta}: It seamlessly integrates MAML and GNNs in a straightforward manner, leveraging the MAML framework to acquire useful prior knowledge from previous tasks during the process of learning node embeddings, enabling it to rapidly adapt to novel tasks.

\noindent \textbf{GPN} \cite{ding2020graph}: It adopts the concept of Protonet for the few-shot node classification task. It uses a GNN-based encoder and evaluator to learn node embeddings and assess the importance of these nodes, while assigning novel samples to their closest categories.

\noindent \textbf{G-Meta} \cite{huang2020graph}: It constructs an individual subgraph for each node, transmits node-specific information within these subgraphs, and employs meta-gradients to learn transferable knowledge based on the MAML framework.

\noindent \textbf{Meta-GPS} \cite{liu2022few}: It cleverly introduces prototype-based parameter initialization, scaling, and shifting transformations to better learn transferable meta-knowledge within the MAML framework and adapts to novel tasks more quickly.

\noindent \textbf{X-FNC} \cite{wang2023few}: It first performs label propagation to obtain rich pseudo-labeled nodes based on Poisson learning, and then filters out irrelevant information through classifying nodes and an information bottleneck-based method to gather meta-knowledge across different meta-tasks with extremely supervised information.

\noindent \textbf{COSMIC} \cite{wang2023contrastive}: It proposes a contrastive meta-learning framework, which first explicitly aligns node embeddings by contrasting two-step optimization within each episode, and then generates hard node classes through a similarity-sensitive mixing strategy.

\noindent \textbf{TLP} \cite{tan2022transductive}: It introduces the concept of transductive linear probing, initially pretraining a graph encoder through graph contrastive learning, and then applying it to obtain node embeddings during the meta-testing phase for downstream tasks. 

\noindent \textbf{TEG} \cite{kim2023task}: It designs a task-equivariant graph few-shot learning framework, leveraging equivariant neural networks to learn adaptive task-specific strategies, aimed at capturing task inductive biases to quickly adapt to unseen tasks.

\subsection{Implementation Details of Baselines}
For traditional meta-learning models, we follow the same settings as \cite{ding2020graph, liu2022few}, and conduct careful hyperparameter search and report their optimal performance. For meta-learning with fewer tasks models, we uniformly use SGC as the graph encoder. Moreover, we adopt the following additional experimental settings. Specifically, for MetaMix, we allow it to perform task augmentation by generating the same number of nodes as those in the original support and query sets for each meta-training task. For MLTI and Meta-Inter, we make them to generate the same number of additional tasks as in our experiments to ensure fairness. For graph meta-learning baselines, we use the hyperparameters recommended in the original papers. All the experiments are conducted by NVIDIA 3090Ti GPUs with the Python 3.7 and PyTorch 1.13 environment.

% \section{Visualization Study}
% \label{visualization_study}
% \begin{figure*}[ht]
% \renewcommand{\thefigure}{S3}
%     \centering
%     \subfigure[within-task mixup]{\includegraphics[width=0.52\textwidth]{picture/intra_task_emb_fin.pdf}}
%     \subfigure[across-task mixup]{\includegraphics[width=0.505\textwidth]{picture/inter_task_embedding_fin.pdf}}
%     \caption{Visualization of the dual-level mixup strategies. In (a), the original nodes in each task are represented by triangles, while the generated nodes are represented by circles, with colors indicating the corresponding classes. In (b), the original tasks are represented by triangles, the generated tasks are represented by circles, and the colors indicate the most similar original tasks.}
%     \label{Vis}
% \end{figure*}
% To visually present the introduced dual-level mixup strategy, we leverage t-SNE \cite{van2008visualizing} to visualize the results of dual-level mixup on the Amazon-clothing dataset under the 5-way 5-shot with 5 tasks few-shot setting, as shown in Fig. \ref{Vis}. Specifically, in the within-task mixup, we randomly select one task consisting of support and query sets. In the across-task, we interpolate 50 tasks, where the task embeddings are the average of the contained node embeddings. According to Fig. \ref{Vis}, we observe that the interpolated nodes within each task and the interpolated tasks generated by SMILE indeed densify the node and task distributions, thereby enhancing the model generalization capability.

\end{document}